\documentclass[usenames,dvipsnames]{article}
\usepackage{iclr2023arxiv,times}

\usepackage{hyperref}
\usepackage{url}
\usepackage{times}
\usepackage{epsfig}
\usepackage{graphicx}
\usepackage{enumitem}

\usepackage{tabularx}
\usepackage{multirow}
\usepackage{booktabs}
\usepackage{caption}
\usepackage{threeparttable}
\usepackage{wrapfig}

\usepackage{textcomp}

\usepackage{color}
\usepackage{breqn}
\makeatletter
\newcommand{\subalign}[1]{%
  \vcenter{%
    \Let@ \restore@math@cr \default@tag
    \baselineskip\fontdimen10 \scriptfont\tw@
    \advance\baselineskip\fontdimen12 \scriptfont\tw@
    \lineskip\thr@@\fontdimen8 \scriptfont\thr@@
    \lineskiplimit\lineskip
    \ialign{\hfil$\m@th\scriptstyle##$&$\m@th\scriptstyle{}##$\hfil\crcr
      #1\crcr
    }%
  }%
}
\makeatother

\usepackage{colortbl}
\definecolor{Orange}{rgb}{0.9,0.5,0}
\definecolor{NavyBlue}{rgb}{0.1, 0.4, 0.8}
\definecolor{Magenta}{rgb}{0.8, 0.1, 0.6}
\definecolor{F7E0D5}{RGB}{247,224,213}
\colorlet{Light}{white!0!F7E0D5}

\hypersetup{colorlinks = true, linkcolor = NavyBlue,
            urlcolor  = gray,
            citecolor = NavyBlue,
            anchorcolor = NavyBlue}
            
\usepackage[capitalize,noabbrev]{cleveref}

\crefformat{section}{\S#2#1#3} 
\crefformat{subsection}{\S#2#1#3}
\crefformat{subsubsection}{\S#2#1#3}
\Crefname{table}{Table}{Tables}
\crefname{table}{Tab.}{Tabs.}
\Crefname{equation}{Equation}{Equations}
\crefname{equation}{Eq.}{Eqs.}
\Crefname{figure}{Figure}{Figures}
\crefname{figure}{Fig.}{Figs.}

\usepackage{xspace}

\usepackage{amsmath}
\usepackage{amsfonts,bm}
\usepackage{bm}

\def\eqref#1{equation~\ref{#1}}

\def\1{\bm{1}}

\DeclareMathAlphabet{\mathsfit}{\encodingdefault}{\sfdefault}{m}{sl}
\SetMathAlphabet{\mathsfit}{bold}{\encodingdefault}{\sfdefault}{bx}{n}

\newcommand{\Pij}{\mathbf{P}_{ij}\xspace}
\renewcommand{\P}{\mathbf{P}\xspace}  

\title{Smooth image-to-image translations with \\  latent space interpolations}

\author{
  Yahui Liu\textsuperscript{1,2},
  Enver Sangineto\textsuperscript{3},
  Yajing Chen\textsuperscript{4},
  Linchao Bao\textsuperscript{5},
  Haoxian Zhang\textsuperscript{5},
  Nicu Sebe\textsuperscript{1},\\
  \textbf{Bruno Lepri\textsuperscript{2} and 
  Marco De Nadai\textsuperscript{2}}\vspace{0.2cm}
\\
  \textsuperscript{1}University of Trento \quad
  \textsuperscript{2}Fondazione Bruno Kessler \quad
  \textsuperscript{3}University of Modena and Reggio Emilia \\
  \textsuperscript{4}Shanghai Jiao Tong University \quad
  \textsuperscript{5}Tencent AI Lab \\
\vspace{-1em}
}

\iclrfinalcopy 
\begin{document}

\maketitle

\begin{abstract}
Multi-domain image-to-image (I2I) translations can transform a source image according to the style of a target domain. One important, desired characteristic of these transformations, is their graduality, which corresponds to a smooth change between the source and the target image when 
their respective latent-space representations are linearly interpolated.
However, state-of-the-art methods usually perform poorly when evaluated using inter-domain interpolations, often producing 
abrupt changes in the  appearance or non-realistic intermediate images.
 In this paper, we argue that one of the main reasons behind this problem is the lack of sufficient inter-domain training data
and
we propose two different regularization methods to alleviate this issue:
 a new shrinkage loss, which compacts the  latent space, 
and a Mixup data-augmentation strategy, which flattens the style representations between domains.
  We also propose a new metric to quantitatively evaluate the degree of the interpolation smoothness, 
 an aspect which is not sufficiently covered by the existing  I2I translation metrics.
Using both our proposed metric and standard evaluation protocols, we show that our regularization techniques can  improve the state-of-the-art 
multi-domain I2I translations by a large margin.
   Our code will be made publicly available upon the acceptance of this article.
\end{abstract}

\vspace{-4mm}
\section{Introduction}
\label{Introduction}

\begin{figure*}[t]	
	\renewcommand{\tabcolsep}{1pt}
	\centering
	\begin{tabular}{c}
	    \includegraphics[width=0.75\linewidth]{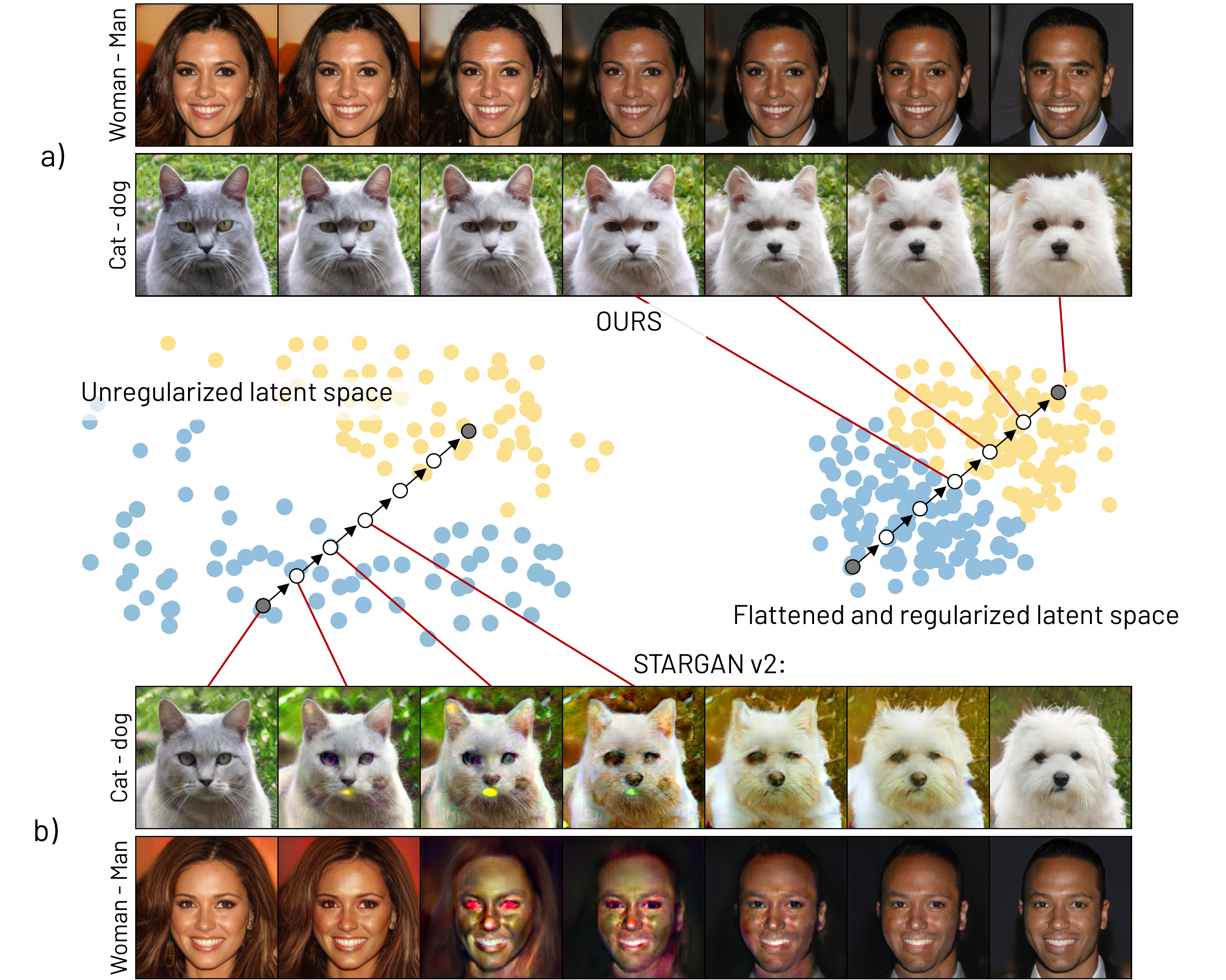}
	\end{tabular}
	\vspace{-0.8em}
	\caption{
	A schematic overview of our regularization approach. 
	In the middle, blue and yellow points represent cat and dog (or woman and men) {\em training samples}, respectively. Linearly interpolating across the two domains leads to traverse a low density area of the true data probability distribution (middle left). Consequently, the corresponding generated images may look unrealistic  (bottom). Conversely, our  regularization  compacts and flattens the latent space  (middle right), resulting in a much smoother
 transition from one generated image to the next  (top). Note that
 the middle figures are  illustrative schemes, but the  bottom and the top figures are images generated using  StarGAN v2 \citep{choi2019stargan}
 and on our regularized space, respectively.
}
	\label{Fig:teaser}
\end{figure*}

The growing interest in generative methods, specifically in image manipulation approaches, goes beyond academia and is also motivated by the enormous application potential, for instance, in the entertainment and fashion industry. 
Modern deep generative networks can artificially change a photo according to some desired ``attribute'' (e.g. changing people's age), and are already applied in leading image editing applications~\citep{Adobe}.
From a scientific point of view, these image transformations are usually called Image-to-Image  (I2I)  ``translations'', and 
the attributes are represented by ``domains'' (e.g., women pictures), where each domain shares some distinctive visual pattern called ``style''.
 In Multi-domain and Multi-modal Unsupervised Image-to-Image Translation (MMUIT), 
a single generator network  maps images into multiple domains, and the process is conditioned by
some random noise in order to   generate diverse images for the same input image
(multi-modal appearance). Moreover, the training dataset is ``unsupervised'', because no image-to-image correspondence is given across the domains.

In this paper, we focus on learning a semantically smooth latent style space, which can be used for continuous MMUIT translations. By linearly interpolating the style representations of the source and the target image, the intermediate generated images should correspond to a gradual transformation of the input image (see Fig~\ref{Fig:teaser} (a)).
Interestingly, while state-of-the-art MMUIT approaches~\citep{choi2019stargan, lee2019drit++} 
can generate highly realistic translations,  
they usually struggle in interpolations across domains. For instance, the across-domain interpolations results of StarGAN-v2
\citep{choi2019stargan} are often unrealistic, with abrupt changes between two close interpolation points
(e.g., see Fig~\ref{Fig:teaser} (b)).
This issue makes it hard to interpolate not to mention extrapolate images, or ``animate'' a translation, and limits the control on the desired {\em degree} of the transformation. 

We argue that one of the main reasons for this problem is the low density of the true data distribution in the inter-domain regions of the latent representation space, which is caused by the lack of sufficient training data representing across-domain images. 
This concept is intuitively shown in Fig~\ref{Fig:teaser}, where the latent-space regions of two domains in two different tasks (cats$\leftrightarrow$dogs and women$\leftrightarrow$men) are densely populated by training samples observed by the generator during training. However, since some real training photos are rare (e.g. people between two genders) or do not exist (e.g. half-cat and half-dog), the inter-domain region has not been sufficiently explored during training. 
Consequently,  when we interpolate between two points belonging to these two domains, the inter-domain area may correspond to meaningless content once decoded by the generator.
A similar phenomenon was studied by \citet{tanielian2020learning},
while \citet{dai2017good} exploit a {\em bad} generator to synthesize fake samples lying in the inter-class regions in a semi-supervised scenario.
Finally, note that the same problem can affect the intra-domain areas: if the domain-specific training samples are too ``scattered'' in a large area,  the generator may overfit the observed training points. 
To solve the overfitting problems related to non-compact representation spaces, Variational Auto Encoders (VAEs)~\citep{kingma2013auto} regularize the latent space using a Kullback-Leibler divergence with respect to an a priori zero-centered Gaussian distribution.
In this paper, we propose an alternative approach which can be used to regularize GAN-based MMUIT networks and produce higher-quality intra and inter-domain interpolations.
Specifically, we propose two simple but effective regularization methods: (1) A new ``shrinkage'' loss for compacting the latent space, and (2) the use of Mixup \citep{verma2019manifold} to generate inter-domain training samples. 

The  shrinkage loss is inspired by the {\em uniform} loss recently proposed by \citet{wang2020understanding} to smooth the latent space of a discriminative network trained using self-supervision. The {\em uniform} loss shares the same goal of the variational regularization in VAEs, that is to make the distributions of the points in the representation space as uniform as possible. However, while the {\em uniform} loss has the effect of (uniformly) {\em spreading} the points on the surface of a unit sphere (using a Gaussian potential kernel), our shrinkage loss forces the points to (uniformly) 
 come {\em closer} to each other. Moreover, we do not need to $L_2$-normalize our representations as in \citet{wang2020understanding}, an operation which is common in self-supervised learning \citep{chen2020simple,byol,caron2020unsupervised} to increase the {\em invariance} of the representations \citep{wang2020understanding} but which can lead to some information loss when the generation of the image details is important.
 
The second regularization method uses a Mixup strategy \citep{zhang2017mixup,verma2019manifold} in the latent space to populate the inter-domain regions with artificially generated training samples. Mixup-based methods are very popular data-augmentation techniques in discriminative networks, and,  recently, they have also been used in a GAN scenario \citep{beckham2019adversarial}.
In our case, we mix the style representations of inter-domain pairs and we use these mixed samples at training time to generate e.g. people between different genders and samples which are not included in the real training data. As far as we know, we are the first proposing a Mixup strategy in an
MMUIT scenario.

Finally, we propose a new metric (Perceptual Proportionality, $P^2$)
to evaluate the semantic smoothness of a latent space. 
As we show in \cref{Evaluation}, $P^2$ is simpler than
 the recently proposed 
Perceptual Path Length (PPL)~\citep{karras2019style} and it avoids different technical problems related to  PPL.  Our contributions can be summarized as follows:
\begin{itemize}[leftmargin=8mm]
    \item We propose a new loss (the {\em shrinkage loss}) and a Mixup-based training strategy to smooth and regularize the latent space of GAN-based MMUIT and TUNIT networks. Both proposals are simple-to-reproduce and can be used in different MMUIT frameworks, jointly with standard losses and different architectural choices.
    \item We show that our approach, when plugged into two state-of-the-art MMUIT and TUNIT frameworks (StarGAN-v2~\citep{choi2019stargan} and \citet{baek2020tunit}) leads to a large boost in the results and it is particularly effective when interpolations are used.
    \item We propose a new metric ($P^2$) that can be used to evaluate the smoothness of a semantic space.
\end{itemize}

\section{Related Work}
\label{Related}

\noindent \textbf{Image-to-image translation}. The goal of I2I translation is to learn a mapping function which changes the domain-specific parts of the source image while keeping the domain-independent part.
Early attempts  are based on paired images~\citep{isola2017image,DBLP:journals/corr/abs-1801-00055,zhu2017toward}, one-to-one domain mappings~\citep{zhu2017unpaired,huang2018multimodal, lee2018diverse,  mao2019mode} and uni-modal deterministic translations~\citep{choi2018stargan, liu2017unsupervised, pumarola2018ganimation}, while recent models focus on MMUIT tasks.
In the latter category, DRIT++~\citep{lee2019drit++} separately models the 
 domain-independent (``content'') and the domain-specific (``style'') image representations
 using a content encoder and a style encoder,
while multi-modal translations are obtained by injecting random noise. 
DMIT~\citep{yu2019multi} adds 
 a domain-specific representation to the content and the style representations of DRIT++. 
StarGAN v2~\citep{choi2019stargan}, the current state-of-the-art method, can generate  high-resolution and diverse images using  a multi-domain discriminator, a style encoder and a noise-to-style mapping network 
(see \cref{sec:problem-formulation}).

Note that in I2I translation ``unsupervised'' means that images are not paired during training. However, they are tagged with domain labels. \citet{baek2020tunit} propose a “Truly UNsupervised” Image Translation (TUNIT) setting, where pseudo-labels are first mined through a clustering procedure and then used for MMUIT tasks.

\noindent \textbf{Latent-space interpolations}. Image interpolations in generative models are obtained using  three main strategies. First, by interpolating the latent-space representations of two different images in VAEs and GANs. For example, in PGGAN~\citep{karras2017progressive} and StyleGAN~\citep{karras2019style, karras2020analyzing}, it is possible to interpolate  two latent codes and generate smooth transitions~\citep{abdal2019image2stylegan, shen2020interpreting, richardson2020encoding, zhu2020domain, abdal2020image2stylegan++}. However, these networks are not designed to translate images in multiple domains. Moreover, linearly travelling  a normally distributed VAE
 latent space  can lead to sub-optimal results~\citep{arvanitidis2018latent}. 

The second strategy  is based on  learning an interpolation function. In HomoGAN, \citet{chen2019homomorphic}  train an interpolation network which interpolates  two latent codes,  at the expense of limited diversity. \citet{shen2020interpreting} identify and exploit the emerging semantics in {\em pretrained} generative models~\citep{karras2017progressive,karras2019style} to linearly traverse the latent space without retraining the networks.

Finally, interpolations can be done using  I2I translations networks (as we do in this paper). 
However, previous MMUIT works  either focus on only  interpolations within a domain  \citep{huang2018multimodal,lee2018diverse}, or they show only qualitative results \citep{lee2019drit++, choi2019stargan}.
In contrast, in this paper we show that our  MMUIT style-space regularization method
can generate realistic and smooth 
 inter-domain interpolations, which  we quantitatively analyze  using both standard MMUIT evaluation protocols and our proposed $P^2$ metric.

\noindent \textbf{Mixup-based regularization}. Mixup~\citep{zhang2017mixup} is a simple yet effective data-augmentation strategy, which is based on blending two input images at the pixel level, and, consequently,  ``blending" also the corresponding image labels.  \citet{verma2019manifold} extend this idea by mixing the  representations in the intermediate layers of the network. 
Importantly, they show that Mixup acts as a latent space regularizer, because it encourages the network to behave linearly between pairs of data points to create smoother class decision boundaries. 
This idea has been used in many discriminative networks \citep{DBLP:conf/iccv/YunHCOYC19,zhang2017mixup,DBLP:conf/nips/SohnBCZZRCKL20} and, recently, ~\citet{beckham2019adversarial} proposed  {\em adversarial Mixup} to regularize the latent space  of an unsupervised auto-encoder.
Our Mixup formulation is inspired by \citet{beckham2019adversarial}, 
which we extend to an MMUIT scenario and of which we propose a multi-domain adaptation.
Note that \citet{beckham2019adversarial} also propose a supervised version of their adversarial Mixup, which, differently from our proposal,  is based on a more complex mixing strategy of the labels, obtained using an ad hoc label embedding function. Moreover, we do not use non-linear mixing strategies of  the samples (e.g., by means of genetic algorithms)  as our goal is to force the information organization in the 
semantic space to be as appropriate as possible under linear interpolations of its elements.

\section{The  Generative Framework}
\label{sec:problem-formulation}

In MMUIT, the training set ($\pmb{\mathcal{X}}$) of real images is  supposed to be composed of $m$ disjoint domains
    ($\pmb{\mathcal{X}} = \bigcup_{k=1}^m \pmb{\mathcal{X}}_k$, 
    $\pmb{\mathcal{X}}_i \cap \pmb{\mathcal{X}}_j = \emptyset, i \neq j$), where each domain  $\pmb{\mathcal{X}}_k$ contains images with the same style. 
In ``truly unsupervised'' I2I translation (TUNIT) \citep{baek2020tunit}, the domain partition is not given but obtained using a  clustering method to  mine a domain (pseudo-)label for each training image.
Thus, without loss of generality, we assume that each  image $\pmb{x} \in \pmb{\mathcal{X}}$ is associated with a label (or a pseudo-label) $y \in \mathcal{Y}$ denoting its domain (i.e.,  $\pmb{x} \in \pmb{\mathcal{X}}_y$).   
    
Our regularization approach (\cref{Regularization}) can be applied to both MMUIT and TUNIT scenarios. In this section, we show the framework for MMUITs, which is mainly inspired by StarGAN v2 \citep{choi2019stargan}, while in \cref{tunitsetting} we show the differences to apply the framework to TUNIT.

Following \citet{choi2019stargan},  the style space $\pmb{\mathcal{S}}$ is \emph{explicitly} modeled
through
 an encoder  $E$ and a  noise-to-style mapping network $F$. 
$\pmb{\mathcal{S}}$ follows the same partition of $\pmb{\mathcal{X}}$: $\pmb{\mathcal{S}} = \bigcup_{k=1}^m \pmb{\mathcal{S}}_k$.
The role of $E$ is to  extract the style code from  an image:  $\pmb{s} = E(\pmb{x})$ 
($\pmb{s} \in \pmb{\mathcal{S}}$). On the other hand,
$F$ (an MLP) is used to inject diversity (appearance ``multi-modality) in the generation process by conditioning with respect to random input noise. 
We sample a random vector ($\pmb{z}\sim\mathcal{N}(\pmb{0},\pmb{I})$) and  we use $F$  to transform $\pmb{z}$ into a style code:
$\pmb{s} = F(\pmb{z})$.
The generator ($G$) translates a source image $\pmb{x}_i \in \pmb{\mathcal{X}}_i$ in a target domain $\pmb{\mathcal{X}}_j$: 
$\hat{\pmb{x}} = G(\pmb{x}_i, \pmb{s}_j)$, where $\pmb{s}_j \in \pmb{\mathcal{S}}_j$ represents the target style which may be either extracted from a  reference image (e.g., $\pmb{s}_j = E(\pmb{x}_j)$) or randomly sampled (e.g., $\pmb{s}_j = F(\pmb{z})$).
This generative framework is trained using different losses, which are briefly described below.

The \textit{style reconstruction} loss 
\citep{huang2018multimodal,zhu2017toward,choi2019stargan} pushes  the target style code  
and the code extracted  from the generated image  to be as close as possible: 
\begin{equation}\label{eq:sty}
    \mathcal{L}_{sty} = \mathbb{E}_{\pmb{x}_i \sim \pmb{\mathcal{X}}_i, \pmb{s}_j \sim\pmb{\mathcal{S}}_j} 
    \left[ \|\pmb{s}_j - E(G(\pmb{x}_i, \pmb{s}_j)) \|_1\right].
\end{equation}
The \textit{diversity sensitive} loss~\citep{choi2019stargan, mao2019mode} is used to generate  diverse images when $G$ is conditioned on  different styles in a same domain:
\begin{equation}\label{eq:ds}
    \mathcal{L}_{ds} = \mathbb{E}_{\pmb{x}\sim\pmb{\mathcal{X}}_i,  \pmb{s}_1, \pmb{s}_2 \sim\pmb{\mathcal{S}}_j} \left[\| G(\pmb{x}, \pmb{s}_1) - G(\pmb{x}, \pmb{s}_2 ) \|_1\right].
\end{equation}
The \textit{cycle consistency} loss~\citep{zhu2017unpaired,choi2018stargan,choi2019stargan}  is used 
to preserve the  content of the source image $\pmb{x}$:
\begin{equation}
\label{eq:cyc_pixelwise}
    \mathcal{L}_{cyc} = \mathbb{E}_{\pmb{x}_i \sim \pmb{\mathcal{X}}_i, \pmb{s}_j \sim\pmb{\mathcal{S}}_j} 
    \left[\|\pmb{x}_i - G(G(\pmb{x}_i, \pmb{s}_j), E(\pmb{x}_i))\|_1\right]
\end{equation}
\noindent
Note that \cref{eq:ds,eq:cyc_pixelwise} work in the pixel space while \cref{eq:sty} is evaluated in the latent style space.

Finally, we adopt the multi-domain discriminator architecture proposed in   StarGAN~\citep{choi2018stargan}. While the discriminator used in \citet{choi2019stargan} requires multiple real/fake binary classification branches, the discriminator of \citet{choi2018stargan} is composed of only two branches, one ($D_{\text{r/f}}$)
 discriminates between real and fake images, and the other branch ($D_{\text{cls}}$) 
estimates a posterior probability over 
$\mathcal{Y}$
 and classifies the domains.
The reason behind this choice will be clarified in \cref{Regularization}. 
The  {\em adversarial} loss is then:
\begin{equation}
\label{eq.adv-loss}
\begin{aligned}
    \mathcal{L}_{\text{adv}} = & \mathbb{E}_{\bm{x}\sim\bm{\mathcal{X}}} [\log D_{\text{r/f}}(\bm{x}) \mathbb{E}_{\bm{x}\sim\bm{\mathcal{X}}_i, \bm{s}\sim\bm{\mathcal{S}}} [\log(1-D_{\text{r/f}}(G(\bm{x}, \bm{s})))],
\end{aligned}
\end{equation}
\noindent
while the {\em domain classification} loss \citep{choi2018stargan} is the cross-entropy loss, which, for the discriminator $D$ and the generator $G$, can be formulated as:
\begin{equation}
\label{eq.dom-cls-loss-D}
    \mathcal{L}_{\text{cls}}^D = \mathbb{E}_{\bm{x}\sim\bm{\mathcal{X}}_i} [\log D_{\text{cls}}(y=i | \bm{x})],
\end{equation}
\begin{equation}
\label{eq.dom-cls-loss-G}
    \mathcal{L}_{\text{cls}}^G = \mathbb{E}_{\bm{x}\sim\bm{\mathcal{X}}_i,  \bm{s}\sim\bm{\mathcal{S}}_j} [\log D_{\text{cls}}(y = j | G(\bm{x},\pmb{s}))].
\end{equation}
We refer the reader to \citep{choi2018stargan,choi2019stargan} and to \cref{additionalarchitecture} for additional details.

\section{Regularizing the  Latent Style Space}
\label{Regularization}

In this section, we introduce our regularization approach, which is based on the {\em shrinkage loss} and on a  Mixup-based sample generation.
Specifically, as mentioned in \cref{Introduction}, the goal of the proposed {\bf shrinkage loss}  is to compact the latent space in order to reduce the regions with a  low true probability density.  This is obtained by:
\begin{equation}
\label{eq.shrinkage}
    \mathcal{L}_{\text{shr}} = \mathbb{E}_{\pmb{s}_1, \pmb{s}_2 \sim \pmb{\mathcal{S}}} [||  \pmb{s}_1 - \pmb{s}_2 ||_2^2 ].
\end{equation}
For each pair of points $(\pmb{s}_1, \pmb{s}_2)$ in the latent space, \cref{eq.shrinkage} penalizes their squared Euclidean distance, in this way fighting against the tendency of $G$ and $D$ to increase the style-space support. 
In  \cref{eq.shrinkage}, the pair $(\pmb{s}_1, \pmb{s}_2)$ 
 is drawn from $\pmb{\mathcal{S}}$
 using a mixed strategy, including both style codes extracted from real images, and randomly generated  codes. More in detail, with probability 0.5, we use two (randomly chosen) real samples
 $\pmb{x}_1 \in \pmb{\mathcal{X}}_i$, $\pmb{x}_2 \in \pmb{\mathcal{X}}_j$, and we extract the corresponding style codes: $\pmb{s}_1 = E(\pmb{x}_1)$, $\pmb{s}_2 = E(\pmb{x}_2)$.
Moreover, with probability 0.5, we use 
$\pmb{z}_1, \pmb{z}_2 \sim\mathcal{N}(\pmb{0},\pmb{I})$ and $\pmb{s}_1 = F(\pmb{z}_1)$, $\pmb{s}_2 = F(\pmb{z}_2)$. 
In practice, we alternate mini-batch iterations in which we use only real samples 
with iterations in which we use only generated samples. 
Note that, in both cases, we may have that both $\pmb{s}_1$ and $\pmb{s}_2$ belong to the same domain:
\cref{eq.shrinkage} is applied to {\em all} the pairwise distances in $\pmb{\mathcal{S}}$ (intra- and inter-domain).
The gradient of $\mathcal{L}_{\text{shr}}$ is directly backpropagated through $E$ and $F$. However, since $G$ directly depends on $\pmb{\mathcal{S}}$, and $D$ indirectly depends on $G$, the effect of $\mathcal{L}_{\text{shr}}$ propagates to the whole generative framework. 

In the {\bf Mixup-based regularization},  inspired by  \citet{beckham2019adversarial}, we use latent-space interpolations to generate 
``fake'' samples to fool the discriminator of the adversarial loss (\cref{eq.adv-loss}). 
Moreover, we 
extend  the domain classification loss  (\cref{eq.dom-cls-loss-G}) to classify mixed samples.
Let:
\begin{equation}
     \text{Mix}(\bm{s}_1, \bm{s}_2, \alpha) = (1-\alpha)\bm{s}_1 + \alpha\bm{s}_2,
\end{equation}
\noindent
where $\bm{s}_1, \bm{s}_2 \sim \pmb{\mathcal{S}}$ and $\alpha$ ($\alpha \in [0, 1]$), as suggested in \citet{verma2019manifold},
is drawn from a Beta distribution: $\alpha \sim \texttt{Beta}(b, b)$. We use $b = 2$, which corresponds to a ``hill'' shape, with most of the mass in the center of the interpolation line. Thereby, most of the mixed samples are generated far from the real data points, i.e., in those areas of 
$\pmb{\mathcal{S}}$ corresponding to a low-density of the true data distribution.  The {\em adversarial mixup} 
loss is:
\begin{equation}
\label{eq.adv-mixup}
\begin{aligned}
    \mathcal{L}_{\text{adv}}^{\text{mix}} = & \mathbb{E}_{\pmb{x}_i \sim\pmb{\mathcal{X}}_i, \pmb{s}_j \sim \pmb{\mathcal{S}}, \alpha \sim \texttt{Beta}(b, b)}[
     \log(1-D_{\text{r/f}}(G(\pmb{x}_i,\bm{s}_{\text{mix}}))],
\end{aligned}
\end{equation}
\noindent
where: $\bm{s}_{\text{mix}}  = \text{Mix}(\bm{s}_i, \bm{s}_j, \alpha)$ and $\pmb{s}_i = E(\pmb{x}_i)$. When sampling $\pmb{s}_j$
($\pmb{s}_j \sim \pmb{\mathcal{S}}$), similarly to \cref{eq.shrinkage},
we use a mixed strategy, alternating: (1) $\pmb{s}_j = F(\bm{z}), \bm{z} \sim\mathcal{N}(\pmb{0},\pmb{I})$ with (2) $\bm{s}_j = E(\bm{x}_j)$, where $\bm{x}_j$ is a  real image different from $\bm{x}_i$ and randomly sampled from the whole training set ($\bm{x}_j \sim \pmb{\mathcal{X}}$). Note that we may have $\bm{s}_i, \bm{s}_j \in \pmb{\mathcal{S}}_i$. 
In \cref{eq.adv-mixup}, $G$   generates an image using the mixed style code ($G(\pmb{x}_i,\bm{s}_{\text{mix}})$) that is used to ``fool'' $D_{\text{r/f}}$. Intuitively, this helps to disentangle $\pmb{\mathcal{S}}$, because unrealistic images,  lying in between $G(\pmb{x}_i, \pmb{s}_i)$
and $G(\pmb{x}_i, \pmb{s}_j)$, are ``moved away'' from the interpolation segment whose endpoints are $\bm{s}_i$ and $\bm{s}_j$.

Analogously to \cref{eq.adv-mixup},
we extend \cref{eq.dom-cls-loss-G} using our {\em domain-mixup classification} loss:
\begin{equation}
\label{eq.domain-mixup}
\begin{aligned}
    \mathcal{L}_{\text{cls}}^{\text{mix}} = & \mathbb{E}_{\subalign{\pmb{x}_i \sim \pmb{\mathcal{X}}_i, \pmb{s}_j \sim\pmb{\mathcal{S}}, i \neq j,\\\alpha \sim \texttt{Beta}(b, b)}}
    [ (1-\alpha) \log D_{\text{cls}}(y = i | G(\pmb{x}_i, \bm{s}_{\text{mix}}) + \alpha \log D_{cls}(y = j | G(\pmb{x}_i, \bm{s}_{\text{mix}}))],
\end{aligned}
\end{equation}


\noindent
where, similarly to \cref{eq.adv-mixup},
$\bm{s}_{\text{mix}}  = \text{Mix}(E(\pmb{x}_i), \pmb{s}_j, \alpha)$
and either $\pmb{s}_j = F(\bm{z}), \bm{z} \sim\mathcal{N}(\pmb{0},\pmb{I})$, or $\bm{s}_j = E(\bm{x}_j)$. 
The constraint  $i \neq j$ is used because
we want to interpolate between samples of different domains
(when $i = j$, then \cref{eq.domain-mixup} corresponds to computing  
\cref{eq.dom-cls-loss-G} twice). 
In \cref{eq.domain-mixup}, we use the binary cross-entropy and the mixing coefficient $\alpha$ is interpreted as a probability value.
Specifically, we want that an image $\pmb{x}_i$, when transformed using the mixed style code $\bm{s}_{\text{mix}}$ ($G(\pmb{x}_i, \bm{s}_{\text{mix}})$),
should belong to domain $\pmb{\mathcal{X}}_i$ with probability $(1-\alpha)$ and to domain $\pmb{\mathcal{X}}_j$ with probability $\alpha$.

Finally, as mentioned in \cref{sec:problem-formulation},
the choice of the StarGAN-like  multi-task discriminator \citep{choi2018stargan} is related to the posterior probability over 
$\mathcal{Y}$ computed by
$D_{\text{cls}}$ and used
in \cref{eq.domain-mixup}. Although \cref{eq.domain-mixup} may be 
adapted to the  multiple independent real/fake  binary  classification branches of StarGAN v2 \citep{choi2019stargan}, the above formulation is more natural. 
\section{Evaluation Protocols}
\label{Evaluation}

\noindent\textbf{Quality and diversity.} We evaluate both the visual quality and the diversity of the generated images. More details in \cref{supp:evaluationprotocol}.

\begin{wrapfigure}[13]{r}{7.8cm}
\vspace{-4mm}
\includegraphics[keepaspectratio,width=7.8cm]{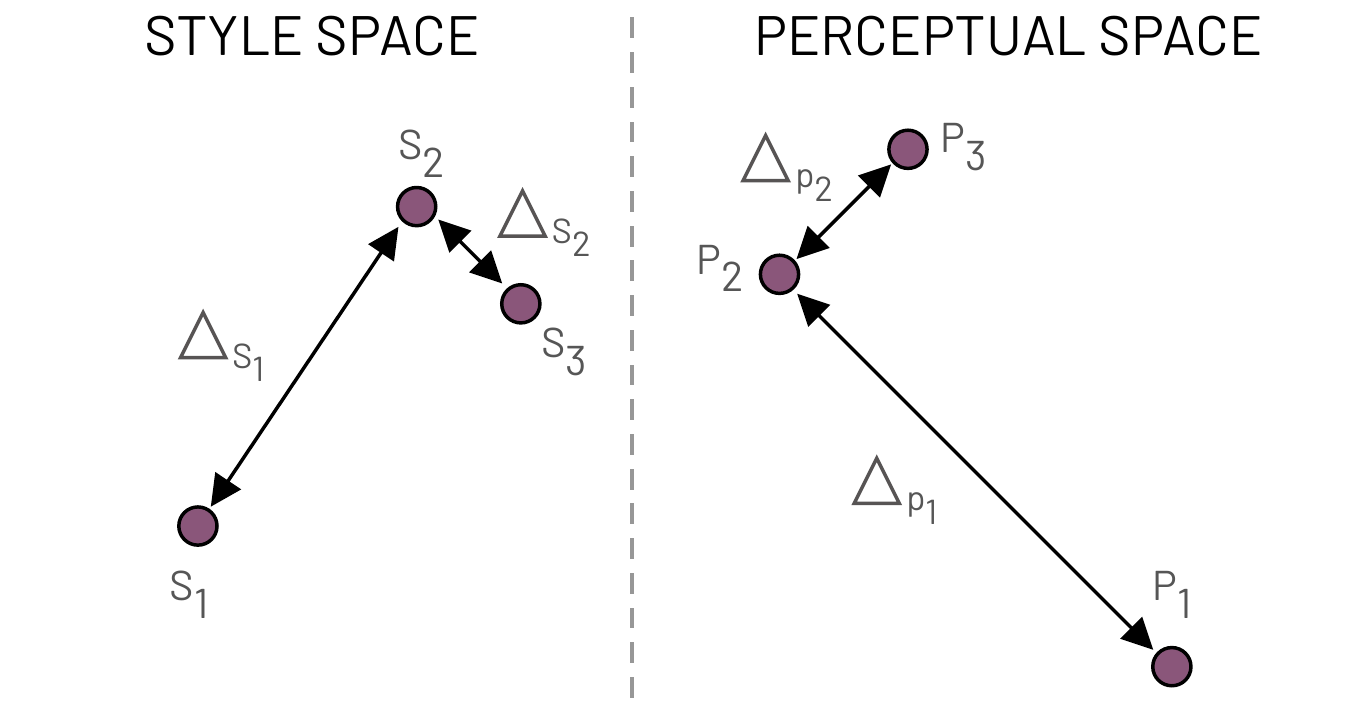}
\vspace*{-6mm}
\caption{A schematic illustration of the  $P^2$ metric.}
    \label{fig:proportionality}
    \vspace{4mm}
\end{wrapfigure} 

\noindent\textbf{Semantic smoothness.} 
\citet{karras2019style} recently proposed the Perceptual Path Length (PPL) to measure the smoothness of a semantic space.
This metric is based on computing the perceptual variation between pairs of generated images under small perturbations ($\epsilon$) in the latent space. The perceptual variation is estimated using an externally trained network, the same used in  LPIPS. However, there are different problems with PPL. First, the  value of $\epsilon$ should be manually estimated depending on the  scale of the latent space, and  PPL decreases quadratically with respect to $\epsilon$. 
Then, PPL can be minimized by a ``collapsed'' generator with no diversity (e.g., constantly generating the same image, independently of the input style code).
Alternative formulations, such as computing the standard deviation of perceptual distances over the interpolation line also suffer from similar problems (e.g., 
perceptual distances between adjacent interpolation points may be highly non-normally distributed).

To solve these issues, we propose a new  smoothness metric called Perceptual Proportionality ($P^2$), whose intuitive idea is shown in 
\cref{fig:proportionality}. The right part of \cref{fig:proportionality} shows a ``perceptual'' space ($\pmb{\mathcal{P}}$), which in practice is the representation space of an externally pretrained network ($\phi$). Specifically, we use the same network used by both the LPIPS and the PPL metric to compute their perceptual distances,  which have been shown to be well aligned with the human perceptual similarity \citep{zhang2018unreasonable}.
On the left of the same figure, we have the style space ($\pmb{\mathcal{S}}$) of the MMUIT  framework we want to evaluate.
Given 3 points on $\pmb{\mathcal{S}}$
($\pmb{s}_1$, $\pmb{s}_2$, $\pmb{s}_3$), we can generate 3 corresponding images which are projected  onto $\pmb{\mathcal{P}}$ using $\phi$
($\pmb{p}_1$, $\pmb{p}_2$, $\pmb{p}_3$). While the absolute distances between these points in the two spaces are different, in an ideal situation, we would like to have the same ratio of their distances. For instance, assuming that: $\Delta_{s1} = ||\pmb{s}_1 - \pmb{s}_2||$, $\Delta_{s2} = ||\pmb{s}_2 - \pmb{s}_3||$
and $\Delta_{p1} = ||\pmb{p}_1 - \pmb{p}_2||$, $\Delta_{p2} = ||\pmb{p}_2 - \pmb{p}_3||$, then, ideally, we should have:
\begin{equation}
\label{eq:ideal-proportion}
\Delta_{p1}/\Delta_{p2}  = \Delta_{s1}/\Delta_{s2}. 
\end{equation}
\noindent
Note that the ratios in \cref{eq:ideal-proportion} are unitless, so they can be compared to each other.
In practice, however, the information organization in $\pmb{\mathcal{P}}$ and in $\pmb{\mathcal{S}}$ will not be exactly the same. Thus, our metric is based on averaging the total errors in \cref{eq:ideal-proportion} computed over a set of triplets of points. In more detail, we sample 3 points ($\pmb{s}_1$, $\pmb{s}_2$, $\pmb{s}_3$)  in 
$\pmb{\mathcal{S}}$ and a source image $\pmb{x} \sim \pmb{\mathcal{X}}$. Then we 
``translate'' $\pmb{x}$ using $\pmb{s}_1$, $\pmb{s}_2$, $\pmb{s}_3$, and we project the generated images onto $\pmb{\mathcal{P}}$,
obtaining: $\pmb{p}_i = \phi(G(\pmb{x}, \pmb{s}_i))$ ($i \in \{ 1,2,3 \}$).
We compute $\Delta_{pj}$ and $\Delta_{sj}$ ($j \in \{ 1,2 \}$) as above, and finally we have:
\begin{equation}
    P^2 = \mathbb{E}_{\pmb{x} \sim \pmb{\mathcal{X}}, \pmb{s}_1, \pmb{s}_2, \pmb{s}_3 \sim \pmb{\mathcal{S}}}
[| \frac{\Delta_{p1}}{\Delta_{p2} + \epsilon} - \frac{\Delta_{s1}}{\Delta_{s2} + \epsilon} |],
\end{equation}
\noindent
where $\epsilon$ is used for numerical stability. 
The lower the value of $P^2$, the more linear is $\pmb{\mathcal{S}}$ with respect to $\pmb{\mathcal{P}}$.
$P^2$ has several advantages: (1) it is simple and relatively fast to compute; (2) it is parameter-free; (3) a 
``mono-modal'' generator that generates always the same image or with a low diversity of outputs, results in a high value of 
$P^2$ (if $\pmb{p}_i \sim \pmb{p}_j$, $i \neq j$, then $\Delta_{p1}/\Delta_{p2} \sim 1$, while $\Delta_{s1}/\Delta_{s2} \neq 1$).

\begin{figure*}[!ht]
	\renewcommand{\tabcolsep}{1pt}
	\renewcommand{\arraystretch}{0.8}
	\centering
	\footnotesize
	\begin{tabular}{ccc}
		\includegraphics[width=0.9\linewidth]{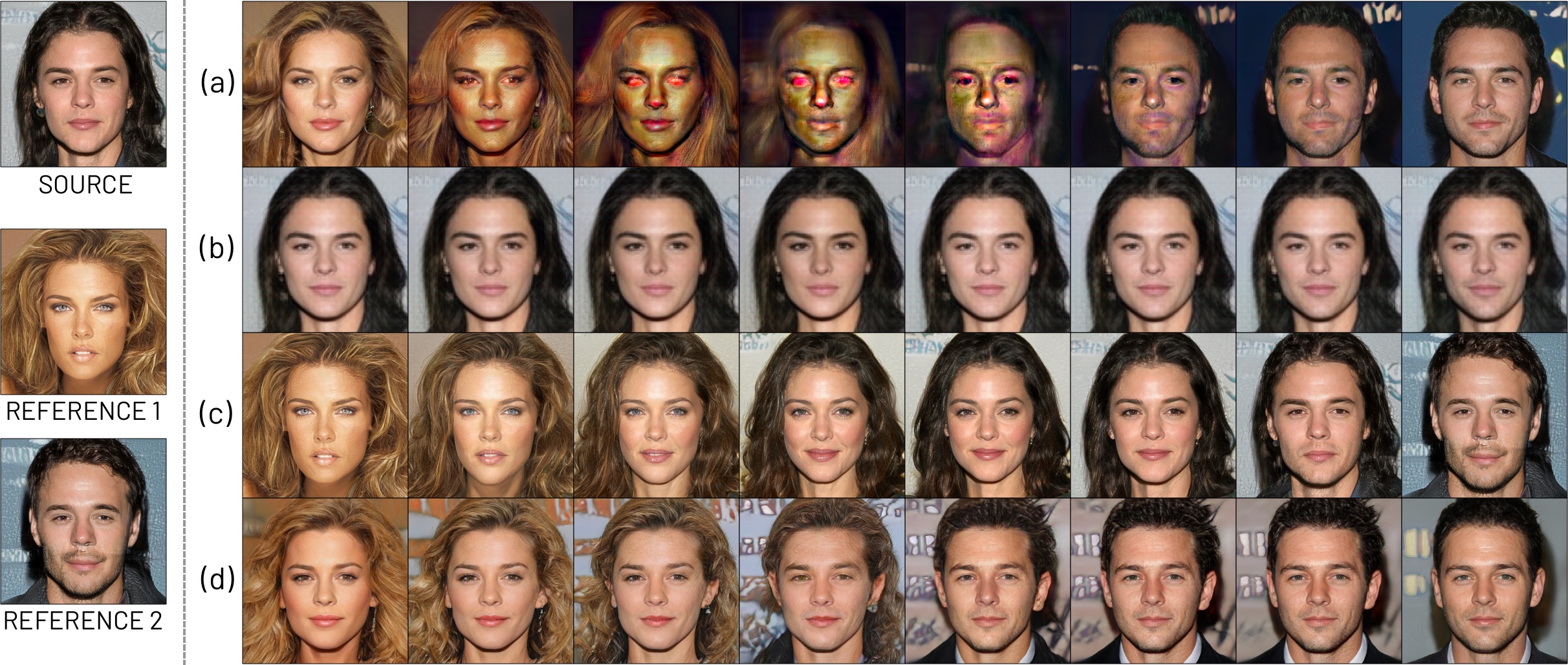}\\
	\end{tabular}
	\vspace{-0.8em}
	\caption{Inter-domain interpolations between genders using CelebA-HQ: (a) StarGAN v2, (b) HomoGAN, (c) InterFaceGAN, (d) our method. All the models use the same source and reference images. Our model generates smoother results while better preserving the source-person identity. 
	}
	\label{Fig:interpolation-celebAHQ}
\end{figure*}

\section{Experiments}
\label{Experiments}

\noindent\textbf{Baselines.} 
We compare our method with state-of-the-art MMUIT, TUNIT and interpolation function learning
approaches (see \cref{Related}). As a representative of the above categories, we use
StarGAN v2~\citep{choi2019stargan}, TUNIT~\citep{baek2020tunit} and HomoGAN~\citep{chen2019homomorphic}, respectively.
Moreover, in the CelebA-HQ experiments, we use also InterFaceGAN~\citep{shen2020interpreting} as a reference for a high-quality generation. Despite this model is not specifically designed for MMUIT and it is based on the {\em very training-intensive} model StyleGAN~\citep{ karras2017progressive, karras2019style}, it performs high-resolution linear interpolations images. All the models are tested using the official source codes.

\noindent\textbf{Datasets and settings.} Following the settings used in StarGAN v2~\citep{choi2019stargan}, we test our method with high-quality images of human and animal faces through CelebA-HQ~\citep{karras2017progressive} and AFHQ~\citep{choi2019stargan}, respectively. We use CelebA-HQ with the gender (\emph{male} and \emph{female}) and the smile (\emph{no smile}, \emph{smile}) domains, while, in AFHQ, we use the {\em cat}, the {\em dog} and the {\em wildlife} domains. We do not use any additional information but the domain labels for the MMUIT setting, while no label is used in the TUNIT setting.
All the images have a $256 \times 256$ resolution. For a fair comparison, we use the same training and testing images for  all the models in each setting. 

\begin{wraptable}{r}{7.5cm}
	\vspace{-0.6em}
\caption{An ablation study of our regularization  losses  using CelebA-HQ with gender translations.}
	\vspace{-0.6em}
\label{tab:ablation}
\small
\begin{tabular}{@{}ll rrrr@{}}
    \toprule
    & \textbf{Model} & \textbf{FID}$\downarrow$ & 
	\textbf{LPIPS}$\uparrow$ &
	\textbf{PPL}$\downarrow$ &
	$\mathbf{P^2}$$\downarrow$  \\
    \midrule
        A: & StarGAN v2 & 42.32 & .443 & 59.25 & .213 \\
        B: & A + $\mathcal{L}_{\text{shr}}$ & 34.44 & .448 & \textbf{27.93} & .277 \\
        C: & A + $\mathcal{L}_{\text{mix}}$ & 26.44 & .245 & 32.95 & \textbf{.173}  \\
        D: & A + $\mathcal{L}_{\text{mix}}$ + $\mathcal{L}_{\text{shr}}$ & \textbf{23.03} & \textbf{.511} & 37.80 & .181 \\
        \bottomrule
    \end{tabular}
\end{wraptable}

\subsection{Ablation study}

In this section, we evaluate the impact of all the  components of our method using  FID, LPIPS and our proposed metric $P^2$. For completeness, we also show the PPL scores (\cref{Evaluation}). 
The results are shown in Tab.~\ref{tab:ablation}, where we separately analyse the contribution of the two proposed regularization methods,
the shrinkage loss $\mathcal{L}_{\text{shr}}$ (\cref{eq.shrinkage}) and the sum of the two Mixup-based losses 
(\cref{eq.adv-mixup} and \cref{eq.domain-mixup}), cumulatively called $\mathcal{L}_{\text{mix}}$ for brevity.
As the starting baseline we use StarGAN v2 \citep{choi2019stargan}, because our losses are added to this method  using exactly its network architectural details and basic training losses (\cref{sec:problem-formulation}). Note that, as mentioned in \cref{sec:problem-formulation}, we use a differently branched discriminator with respect to StarGAN v2 which,
according to \citet{choi2019stargan}, leads to a slightly worse average performance.

 \cref{tab:ablation} (B) shows a relative improvement in all the metrics except $P^2$ with respect to the base model,
confirming the importance of  a compact semantic space to improve the image quality and the diversity of the I2I translations. 
Comparing Tab.~\ref{tab:ablation} (B)  with Tab.~\ref{tab:ablation} (C), we observe  that $\mathcal{L}_{\text{mix}}$ obtains an even higher  improvement on the image quality 
(FID: $-37.52\%$)  and the smoothness degree ($P^2$: $-18.78\%$) with respect to StarGAN v2, at the expense, however, of diversity (LPIPS). Finally, Tab.~\ref{tab:ablation} (D) shows that the combination of mixup and the shrinkage loss  drastically improves both FID and LPIPS with respect to both the ablated methods.
However, the latent-space smoothness degree of the full model is not the best over the tested combinations (e.g., it underperforms 
Tab.~\ref{tab:ablation} (C) when measured with both $P^2$ and PPL).
 We speculate this result might be a consequence of a trade-off in MMUIT models between diversity and smoothness. The higher the diversity of the translations, the more challenging is  to keep   gradual the changes 
  between neighbouring points in the latent space.

\subsection{Comparison with the state of the art}

\begin{figure}[ht]
	\renewcommand{\tabcolsep}{1pt}
	\renewcommand{\arraystretch}{0.8}
	\centering
	\footnotesize
	\begin{tabular}{ccc}
		\includegraphics[width=\linewidth]{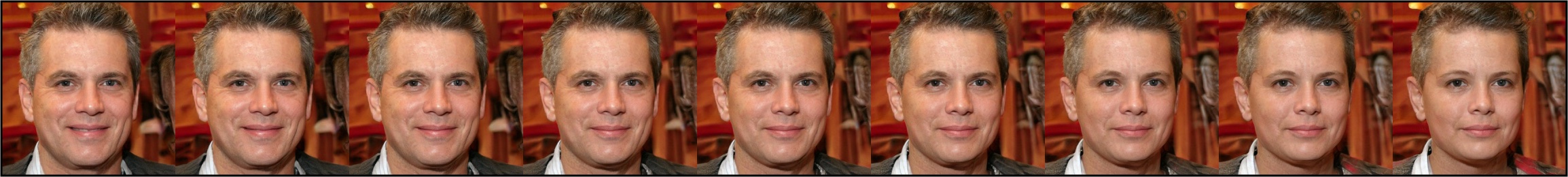}\\
	\end{tabular}
	\vspace{-0.8em}
	\caption{Inter-domain interpolations between multiple domains (gender and expression).}
	\label{fig:multi}
\end{figure}

{\bf Qualitative comparison.}
We first compare our method with state-of-the-art approaches on CelebA-HQ. As shown in \cref{Fig:interpolation-celebAHQ}, the images are obtained by linearly interpolating the style codes between $\pmb{s}_1 = E(\pmb{x}_1)$ and $\pmb{s}_T = E(\pmb{x}_2)$, where 
$\pmb{x}_1$ and $\pmb{x}_2$ are two {\em reference} images belonging to two different domains. The intermediate style codes ($\{ \pmb{s}_1, \ldots, \pmb{s}_T \}$)
are used to transform a common
 {\em source} image $\pmb{x}$, leading to   a set of images $\{ G(\pmb{x}, \pmb{s}_1), \ldots, G(\pmb{x}, \pmb{s}_T) \}$ for each compared generation method, which are shown in the corresponding rows of \cref{Fig:interpolation-celebAHQ}.
Specifically, \cref{Fig:interpolation-celebAHQ} (a) shows that StarGAN v2 generates artifacts and unrealistic results, especially in the center of the interpolation line between the two  domains.
On the other hand, the interpolation results of HomoGAN are very smooth, since neighbouring images are almost indistinguishable the one from the other (\cref{Fig:interpolation-celebAHQ} (b)). However, the HomoGAN translations endpoints ($G(\pmb{x}, \pmb{s}_1)$ and $G(\pmb{x}, \pmb{s}_T)$) change very little the one from the other, calling into question whether the model can do image-to-image translations. 
Conversely, our method (\cref{Fig:interpolation-celebAHQ} (d)), successfully translates the source image into the target domains and
the intermediate translation results are both highly  
 realistic and gradually changing.
The quality of our results is comparable with the reference model InterFaceGAN (\cref{Fig:interpolation-celebAHQ} (c)), which is based on the   computationally very intensive training of StyleGAN with high-resolution images~\citep{karras2019style, karras2020analyzing, karras2017progressive}. \cref{fig:multi} we show an example where we perform inter-domain interpolations between multiple domains at the same time. We show additional qualitative results in \cref{AdditionalQualitative}.

\begin{figure}[ht]
	\renewcommand{\tabcolsep}{1pt}
	\renewcommand{\arraystretch}{0.8}
	\centering
	\footnotesize
	\begin{tabular}{ccc}
		\includegraphics[width=\linewidth]{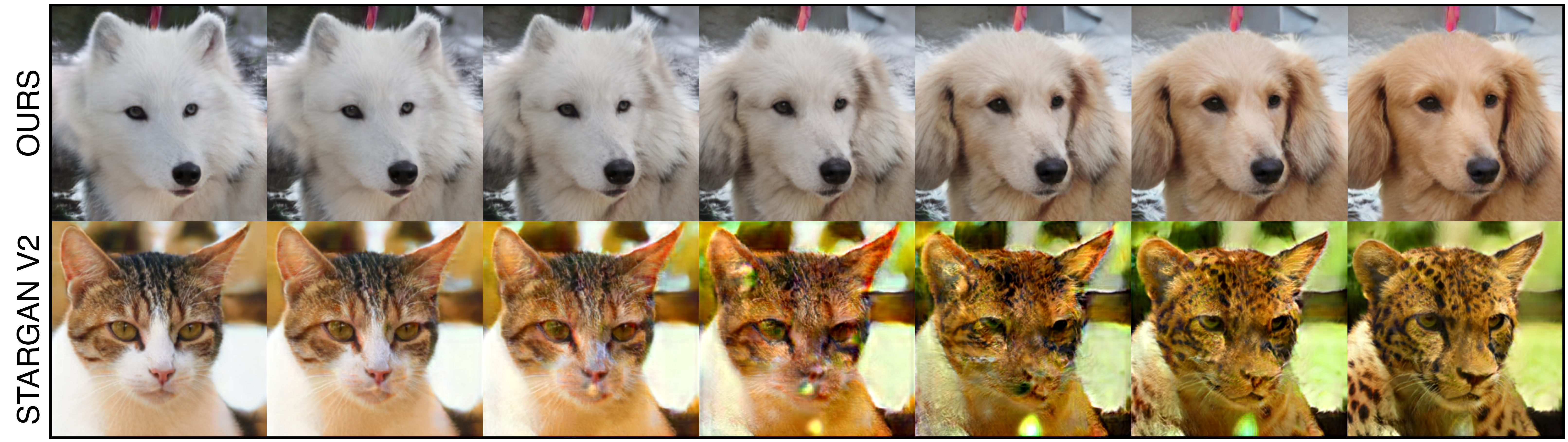}\\
	\end{tabular}
	\vspace{-0.8em}
	\caption{Inter-domain interpolations between StarGAN v2 and our model in  on AFHQ.
	}
	\label{Fig:interpolation-AFHQ}
\end{figure}

\cref{Fig:interpolation-AFHQ} shows the results on AFHQ. We observe that our model interpolates animal images very smoothly, generating inter-species animals. Conversely, StarGAN v2 interpolations contain abrupt changes,  artifacts and unrealistic results, similarly to those generated  with CelebA-HQ images. 
Note that we cannot use InterFaceGAN on this dataset due of the lack of a publicly available pretrained StyleGAN model on AFHQ.
In the \cref{AdditionalQualitative} we show additional  comparative results. 

{\bf Quantitative comparison.}
In Tab.~\ref{tab:celebahq-quality}, we use the
 CelebA-HQ dataset and we quantitatively compare our method with the other approaches with respect to the 
 image quality (FID) and  diversity (LPIPS). As expected,  InterFaceGAN achieves the best FID, but with a very low diversity degree (LPIPS scores).
 In fact, style and content are not disentangled in the StyleGAN latent space, and this prevents the use of a noise-to-style mapping network (similar to our $F$, see \cref{sec:problem-formulation}) to inject style-specific diversity in the image translations.
The  quantitative results of HomoGAN confirm its qualitative evaluation,
with a diversity degree  even lower than InterFaceGAN.
StarGAN v2 clearly outperforms HomoGAN, and  our method largely outperforms StarGAN v2 with respect to  all the metrics.

\begin{table}[!ht]
\caption{Image quality (FID) and translation diversity (LPIPS) measured on the CelebA-HQ dataset. \textsuperscript{\S}Reference: StyleGAN-based model with 1024$\times$1024 images.\textsuperscript{$\dagger$}: always generates the same image.}
	\vspace{-0.6em}
	\label{tab:celebahq-quality}
	\label{tab:celebahq-PIR}
    \setlength{\tabcolsep}{3pt}
    \small
	\centering
	\begin{threeparttable}
	\begin{tabularx}{\columnwidth}{@{}Xrr rr rr rr rr@{}}
	\toprule
	\multirow{2}{*}{\textbf{Model}} & \multicolumn{2}{c}{\textbf{FID}$\downarrow$} &  \multicolumn{2}{c}{\textbf{LPIPS}$\uparrow$} &  \multicolumn{2}{c}{\textbf{PPL}$\downarrow$} & \multicolumn{2}{c}{$\mathbf{P^2}$$\downarrow$}
	\\ \cmidrule(lr){2-3} \cmidrule(lr){4-5} 
	\cmidrule(lr){6-7} \cmidrule(lr){8-9} 
	& Gender & Smile & Gender & Smile & Gender & Smile &  Gender & Smile
	\\ \midrule
	HomoGAN~\citep{chen2019homomorphic} & 55.23 & 58.02 & .001 & $<.001$ & \textbf{5.42}\tnote{$\dagger$} & \textbf{1.17}\tnote{$\dagger$} & .250 & .220 \\ 
	StarGAN v2~\citep{choi2019stargan} & 42.32 & 28.16 & .443 & .413 & 59.25 & 40.79 & .213 & .178 \\ 
	Ours & \textbf{23.03} & \textbf{22.62} & \textbf{.511} & \textbf{.480} & 37.80 & 35.04 & \textbf{.181} & \textbf{.167}  \\
	\midrule \midrule
	InterFaceGAN~\citep{shen2020interpreting}\tnote{\S} & 13.75 & 12.81 & .067 & .027 & 51.73 & 24.24 & .157 & .123 \\ 
	\bottomrule
	\end{tabularx}
	\end{threeparttable}
	\vspace{-0.5em}
	
\end{table}

We also quantitatively measure the latent-space smoothness  using PPL and our proposed $P^2$  (Tab.~\ref{tab:celebahq-PIR}). 
In Tab.~\ref{tab:celebahq-PIR}, HomoGAN gets the best PPL, which is significantly lower than all other models. However, as previously seen in the qualitative results, this model generates images with very little changes along the interpolation lines. Interestingly, our proposed $P^2$ metrics is more aligned with the qualitative results, since assigns to HomoGAN the lowest-ranking value over the three compared methods (see \cref{Evaluation}).
Compared to StarGAN v2, our approach drastically improves both the PPL and the $P^2$  scores, quantitatively showing that 
our regularization methods can  smooth the semantic-space representations. 
We note that InterfaceGAN gets  better $P^2$ than our model. However,  StyleGAN (the model on which InterfaceGAN is based)
 is massively trained to disentangle  the  factors of variation of its semantic space \citep{karras2019style, karras2020analyzing}. In $\cref{sec:evaluationp2}$ we show the evaluation of $P^2$.

 \begin{wraptable}{r}{7.5cm}
	\vspace{-0.8em}
\caption{Quantitative evaluation on AFHQ. }
	\vspace{-0.6em}
	\label{tab:afhq}
    \renewcommand{\tabcolsep}{3pt}
    \renewcommand{\arraystretch}{1}
\small
\begin{tabular}{@{}lc rrrrr@{}}
	\toprule
	\textbf{Model} & \textbf{Setting} & \textbf{FID}$\downarrow$ & \textbf{LPIPS}$\uparrow$ & \textbf{PPL}$\downarrow$ & 
	$\mathbf{P^2}$$\downarrow$ \\
	\midrule
	StarGAN v2 & \multirow{2}{*}{MMUIT} & 15.64 & .435 & 79.62 & .226 \\
     Ours & & \textbf{11.56} & \textbf{.454} &  \textbf{19.49} & \textbf{.211} \\
	\midrule 
	\citet{baek2020tunit} & \multirow{2}{*}{TUNIT} & 19.67 & \textbf{.442} & 22.80 & .173  \\
     Ours & & \textbf{17.23}  & .307  & \textbf{17.94} & \textbf{.148}  \\
	\bottomrule
	\end{tabular}
\vspace*{-2mm}
\end{wraptable}

Tab.~\ref{tab:afhq} shows the quantitative results for the more challenging AFHQ dataset, where there is a more significant inter-domain difference  than in CelebA-HQ.  Our method outperforms with a significant margin all the other tested approaches in all the settings and with all the metrics, except \citet{baek2020tunit} with respect to the LPIPS metric.
Due to the lack of space, we show the qualitative results of the TUNIT setting in \cref{AdditionalQualitative}.
In the AFHQ dataset, we do not include HomoGAN because that model requires well-aligned training images having the same orientation~\citep{HomoGANproblems}
and this makes it hard to train HomoGAN
on the animal face images of AFHQ. 

Overall, the quantitative and the qualitative analysis show that our regularization method drastically improves the state-of-the-art multi-domain translations in both the MMUIT and TUNIT settings.

\section{Conclusion}

In this paper, we presented a  regularization approach for MMUIT networks which is based on the hypothesis that the true data distribution in the inter-domain regions of the representation space is not well modeled because of the inherent scarcity of inter-domain training data. To solve this problem, we propose two simple, yet very effective regularization approaches, respectively based on the shrinkage loss (which compacts the latent space) and on a Mixup data augmentation strategy (which populates the regions across two domains). 
Moreover, we propose a new metric to explicitly evaluate the semantic smoothness of a style space.

Using both our $P^2$ metric and common MMUIT evaluation protocols, we  
showed that
the proposed regularization losses  can be plugged in existing MMUIT frameworks,  leading to a significant quality improvement of the results in all the tested  MMUIT settings.

\bibliographystyle{iclr2023_conference}
\bibliography{egbib}

\appendix

\section{Our framework in the TUNIT Setting}
\label{tunitsetting}

In this section, we describe the generative framework we adopted for the TUNIT setting, which is based on the method presented in \cite{baek2020tunit}. Similarly to the MMUIT setting described in the main paper, in which we modify StarGAN v2 adding our losses to the StarGAN v2 native losses and changing the discriminator, also for the TUNIT setting
we modify the approach proposed in \cite{baek2020tunit}  by:
\begin{enumerate}
    \item adding our losses to the  losses used in \cite{baek2020tunit} and
    \item replacing the discriminator of \cite{baek2020tunit} with our discriminator (the latter being described in Sec. 3 of the main paper and in more detail in Sec.~\ref{sec.discriminator}).
\end{enumerate}

For completeness, we briefly describe below the approach proposed in \cite{baek2020tunit}, emphasizing that this is not our contribution. We believe that the main interest in describing the details of the method proposed by \cite{baek2020tunit} is that their losses are drastically different from those used in StarGAN v2 (e.g., see below the Mutual Information maximization or the contrastive loss). Despite that, as shown in Sec. 6.2 of the main paper and in Sec. \ref{AdditionalQualitative}, 
our regularization methods can successfully  be used jointly with the losses and the architecture proposed in \cite{baek2020tunit}, showing the generality of our regularization proposal.

\subsection{The Adopted TUNIT Framework}

The architecture proposed by \cite{baek2020tunit}  is composed of an encoder network $E$, which has two branches for pseudo-label classification $E_{C}$ and style extraction $E_{S}$, a generator $G$ and a multi-task discriminator $D$, which has as many output branches as the number of domains $m$. 
Since in the TUNIT setting the domain partition is not available, the model jointly learns to cluster the real images and to translate them into different domains.  

\noindent\textbf{Computing the pseudo-labels.} \cite{baek2020tunit} use IIC \citep{ji2019invariant} to cluster the real images in multiple domains and extract the corresponding pseudo-labels. The main idea in IIC is that two  augmented versions of the same image (e.g. obtained using horizontal flipping) should be similarly classified.
For this reason, they define the joint probability matrix $\P \in \mathbb{R}^{m \times m}$:
\begin{equation}
\label{eq:jointprobmat}
\P=\mathbb{E}_{\pmb{x} \sim \bm{\mathcal{X}}}[E_{C}(\pmb{x})\cdot E_{C}(f(\pmb{x}))^T],
\end{equation}
where $f$ is the data augmentation function.
Then, they maximize the Mutual Information (MI)
computed as:

\begin{equation}
\label{eq:iic}
\mathcal{L}_{MI} = \sum_{i=1}^m \sum_{j=1}^m \Pij \ln \frac{\Pij}{\P_i \P_j},
\end{equation}
where $\P_i$ denotes the $m$-dimensional marginal probability vector, and $\Pij$ denotes the joint probability of domain $i$ {\em and} domain $j$. For more details, we refer to \cite{ji2019invariant,baek2020tunit}.

To further help domain classification, \cite{baek2020tunit} use also a contrastive loss~\citep{hadsell2006dimensionality}:
\begin{equation}
\label{eq:contrastive}
\mathcal{L}_{style}^E = \mathbb{E}_{\pmb{x} \sim \bm{\mathcal{X}}}\left[ -\log\frac{\exp(E_{S}(\pmb{x})\cdot {E_{S}(f(\pmb{x}))}/\tau)}{\sum_{i=0}^N \exp(E_{S}(\pmb{x})\cdot E_S(\pmb{x}_i^-)/\tau)}\right],
\end{equation} 
where the sum in the denominator is over $N$ \emph{negative} samples $\pmb{x}^-$ ($\pmb{x}^- \neq \pmb{x}$) contained in a queue $Q$ (we refer to~\cite{he2020momentum} for more details).

\noindent\textbf{Learning to translate images.} \cite{baek2020tunit} force the target style code and the code extracted from the generated image to be as close as possible:
\begin{equation}
\begin{split}
\label{eq:stycontrastive}
\mathcal{L}_{style}^G &= \mathbb{E}_{\pmb{x} \sim \bm{\mathcal{X}}, \pmb{s}\sim \bm{\mathcal{S}}} [ \\
& -\log\frac{\exp(E_{S}(G(\pmb{x}, \pmb{s})) \cdot \pmb{s})}{\sum_{i=0}^N \exp(E_{S}(G(\pmb{x}, \pmb{s})) \cdot E_S(\pmb{x}_i^-)/\tau)} ],
\end{split}
\end{equation}
where $\pmb{s} = E_{S}(\Tilde{\pmb{x}})$ is extracted from a randomly sampled reference image $\Tilde{\pmb{x}} \sim \bm{\mathcal{X}}$, and $\pmb{x}_i^-$ denotes the negative samples as described in Eq.~(\ref{eq:contrastive}).

Then, the \emph{image reconstruction loss} is used to reconstruct the source image. It is defined as:
\begin{equation}
\label{eq:reconstruction}
\mathcal{L}_{rec} = \mathbb{E}_{\pmb{x}\sim \bm{\mathcal{X}}}[\lVert \pmb{x} - G(\pmb{x}, E_{S}(\pmb{x})) \rVert_1].
\end{equation}

Finally,  \cite{baek2020tunit} use an adversarial loss based on a multi-task discriminator which is similar to the StarGAN v2 discriminator~\citep{choi2019stargan} (see Sec. 3 of the main paper).
Conversely, we  adopt the multi-domain discriminator architecture proposed in StarGAN~\citep{choi2018stargan}, analogously  to what we used for the MMUIT setting (see Sec. 3 of the main paper).

As above mentioned, our TUNIT model differs from \cite{baek2020tunit} because of the discriminator and the addition of our regularization losses.

\section{Implementation Details}

\subsection{The Overall Architecture}
\label{additionalarchitecture}
\begin{figure*}[!ht]
	\renewcommand{\tabcolsep}{1pt}
	\renewcommand{\arraystretch}{0.8}
	\centering
	\footnotesize
	\begin{tabular}{ccc}
	\includegraphics[width=0.9\linewidth]{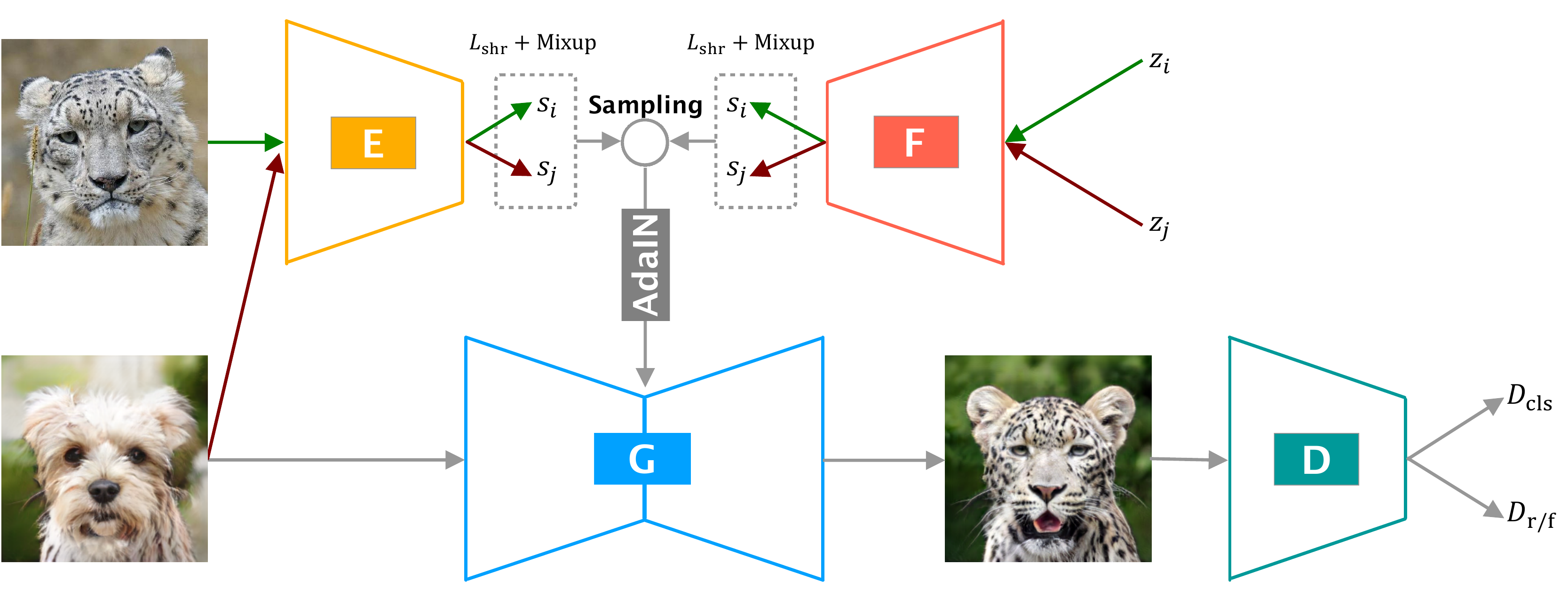}\\
	\end{tabular}
	\caption{In the MMUIT setting, the generator $G$ takes an image $\bm{x}$ and a style code $\bm{s}$ as input and generates an image that is fed to the  discriminator $D$. $D$ learns to classify the images into their own domain ($D_{\text{cls}}$) and to discriminate between real and fake images ($D_{\text{r/f}}$). The encoder $E$ and the noise-to-style mapping network $F$ are two instruments  to get a specific style code $\bm{s}$. Our shrinkage loss $\mathcal{L}_{\text{shr}}$ and the mixup-based losses  regularize the style space.
	}
	\label{Fig:architecture}
\end{figure*}

\cref{Fig:architecture} shows the architecture of our framework in the MMUIT setting. 

\subsection{The Discriminator}
\label{sec.discriminator}

\Cref{tab:dis_architecture} shows the details of the discriminator we used in both the MMUIT and the TUNIT setting.


\begin{table}[h]
\centering
\newcommand{\shape}[6]{$\makebox[\widthof{#1}][c]{#4}\times\makebox[\widthof{#2}][c]{#5}\times\makebox[\widthof{#3}][c]{#6}$}
\begin{tabular}{llcc}
\toprule
& \textsc{Layer} & \textsc{Resample} & \textsc{Output shape}\vspace{0.5mm}\\
\toprule 
& Image $\bm{x}$ & - & \shape{256}{256}{32ch}{256}{256}{3}\\
\midrule
& Conv3$\times$3 & -  & \shape{256}{256}{16ch}{256}{256}{64}\\
& ResBlk & AvgPool & \shape{256}{256}{32ch}{128}{128}{128}\\
& ResBlk & AvgPool & \shape{256}{256}{32ch}{64}{64}{256}\\
& ResBlk & AvgPool & \shape{256}{256}{32ch}{32}{32}{512}\\
& ResBlk & AvgPool & \shape{256}{256}{32ch}{16}{16}{512}\\
& ResBlk & AvgPool & \shape{256}{256}{32ch}{8}{8}{512}\\
& ResBlk & AvgPool & \shape{256}{256}{32ch}{4}{4}{512}\\
\midrule
\multirow{4}{*}{$D_{\text{r/f}}$} & LReLU & - & \shape{256}{256}{32ch}{4}{4}{512}\\
& Conv4$\times$4 & - & \shape{256}{256}{32ch}{1}{1}{512}\\
& LReLU & - & \shape{256}{256}{32ch}{1}{1}{512}\\
& Conv1$\times$1 & - & \shape{256}{256}{32ch}{1}{1}{1}  \\
\midrule
\multirow{4}{*}{$D_{\text{cls}}$} & LReLU & - & \shape{256}{256}{32ch}{4}{4}{512}\\
& Conv4$\times$4 & - & \shape{256}{256}{32ch}{1}{1}{512}\\
& LReLU & - & \shape{256}{256}{32ch}{1}{1}{512}\\
& Conv1$\times$1 & - & \shape{256}{256}{32ch}{1}{1}{$m$}  \\
\bottomrule
\end{tabular}
\caption{The discriminator architecture. $m$ is the number of domains.}
\label{tab:dis_architecture}
\end{table}

\section{$P^2$ Metric}

\subsection{Implementation Details}

\noindent{\bf Perceptual distances.}
As mentioned in the main paper, the proposed $P^2$ metric is based on  distances computed over the  space $\pmb{\mathcal{P}}$. In practice, we compute the involved perceptual distances using an 
 externally pre-trained network ($\phi$), the same  network used by both the 
LPIPS and the PPL metric,   which  
 has been shown to
be well aligned with the human perceptual similarity \citep{zhang2018unreasonable}.
However, although \cite{zhang2018unreasonable} (who first proposed LPIPS)  claim that their distance is a metric, their formulation is based on the squared Euclidean distance between the features of different layers of $\phi$:

\begin{equation}
\label{eq:lpips_org}
    d(\pmb{x}_1, \pmb{x}_2) = \sum_{l}\frac{1}{H_lW_l}\sum_{h,w} w_l \| \phi_l(\pmb{x}_1(h,w)) - \phi_l(\pmb{x}_2(h,w)) \|_2^2,
\end{equation}

\noindent
where $\phi_l(\pmb{x}(h,w))$ is the feature  at position $(h,w)$ in the convolutional feature map of layer $l$, and $w_l$ is a learned layer-specific weight.  Thus, Eq.(\ref{eq:lpips_org}) does not satisfy  the triangle inequality, which is necessary for a distance to be a proper metric. For this reason, we use a slightly different  formula:

\begin{equation}
\label{eq:lpips_ours}
    d'(\pmb{x}_1, \pmb{x}_2) = \sum_{l} \frac{1}{H_lW_l}\sum_{h,w} w_l \| \phi_l(\pmb{x}_1(h,w)) - \phi_l(\pmb{x}_2(h,w)) \|_2.
\end{equation}


In \cite{zhang2018unreasonable}, $\phi$ is an AlexNet~\citep{krizhevsky2017imagenet} pre-trainted on ImageNet, while the weights $\{w_l\}$ are 
trained in order to mimic the human perceptual distance. Accordingly, we have re-trained the weights $\{w_l\}$ following the protocol and the dataset used in \cite{zhang2018unreasonable} (which is {\em different} from the I2I translation datasets used in the main paper), but replacing Eq.\eqref{eq:lpips_org} with Eq.\eqref{eq:lpips_ours}.

Finally, $\Delta_{p1}$ in Eq. (12) of the main paper is computed using: $\Delta_{p1} = d'(\pmb{x}_1, \pmb{x}_2)$ (and similarly for $\Delta_{p2}$).

\noindent{\bf Computing $P^2$ without an explicit style space $\pmb{\mathcal{S}}$.}
In InterFaceGAN~\citep{shen2020interpreting}, there is no separation between the ``content'' and the ``style'' representations, thus we cannot
sample three arbitrary points $\pmb{s}_i$,  in $\pmb{\mathcal{S}}$  and then generate $G(\pmb{x}, \pmb{s}_i)$ ($i \in \{1,2,3\}$)
as in Sec. 5 of the main paper.
For this reason, we approximate the sampling procedure as follows. We ask InterFaceGAN to interpolate between two reference latent codes using $T$ equally spaced interpolation points. In this way we get a sequence of $T$ generated images $I = (\hat{\pmb{x}}_1, ..., \hat{\pmb{x}}_T)$. Then 
we randomly choose $i,k \in \{1, ..., T \}$ and we select 
$\hat{\pmb{x}}_i$, $\hat{\pmb{x}}_{i+k}$ and $\hat{\pmb{x}}_{i+2k}$ in $I$. In this way the three chosen
 images are selected using a constant step ($k$). As a consequence, $\Delta_{s1} = \Delta_{s2}$ and $\Delta_{s1}/\Delta_{s2} = 1$. Hence, Eq. (12) in the main paper can be rewritten as:
 
\begin{equation}
    P^2 = \mathbb{E}_{\pmb{x} \sim \pmb{\mathcal{X}}, i, k \sim \{1,...,T\}}
[| \frac{\Delta_{p1}}{\Delta_{p2} + \epsilon} - 1 |],
\end{equation}

\noindent
where  $\Delta_{p1} = d'(\hat{\pmb{x}}_i, \hat{\pmb{x}}_{i+k})$ and  $\Delta_{p2} = d'(\hat{\pmb{x}}_{i+k}, \hat{\pmb{x}}_{i+2k})$.
For a fair comparison, we adopt this procedure for all the tested methods (including ours).

\subsection{Evaluation}
\label{sec:evaluationp2}
\begin{figure}[!ht]%
    \centering%
    \includegraphics[width=\columnwidth]{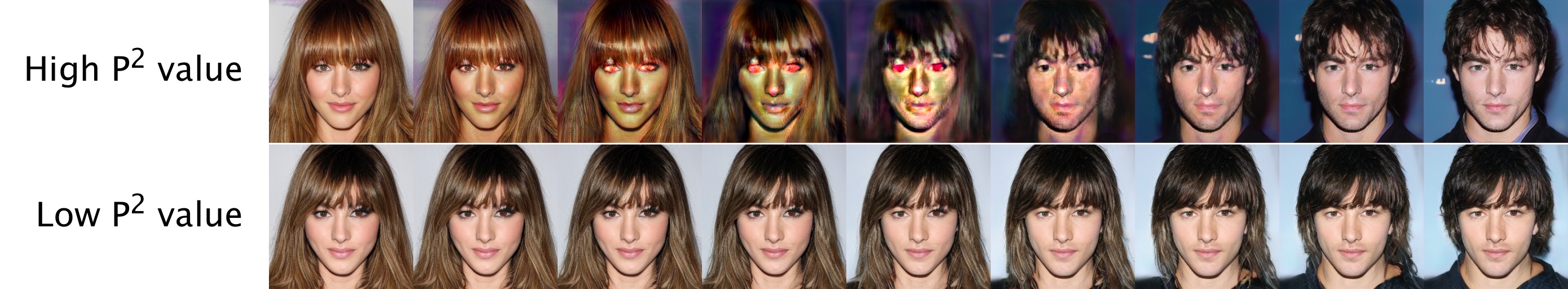}%
    \caption{Random examples with low $P^2$ ($\le$ 10th percentile) in the first row, while in the second row we show some examples with high $P^2$ ($\ge$ 90th percentile). There is a clear correlation between $P^2$ scores and the smoothness of interpolations.}
    \label{fig:comparePP}%
\end{figure}

To evaluate , we use the PPL evaluation protocol adopted in StyleGAN~\citep{karras2020analyzing},  we used $P^2$ to rank the of the per-image $P^2$ score interpolations. \cref{fig:comparePP} shows the top most row shows the interpolation with the highest $P^2$ value, while the bottom row corresponds to the lowest score. 
\\


\section{Evaluation Protocol}
\label{supp:evaluationprotocol}

\subsection{FID-computation   Details}

The FID scores are computed using the interpolation results as follows. For each $\bm{\mathcal{X}}_i\rightarrow \bm{\mathcal{X}}_j$ domain translation, we use 1,000 test source images. For each source image ($\pmb{x}$), we separately randomly select two different reference images ($\bm{x}_1 \in \bm{\mathcal{X}}_i$ and $\bm{x}_2 \in \bm{\mathcal{X}}_j$), which are used to extract  the start and the end 
style codes ($\bm{s}_1 = E(\bm{x}_1) \in \bm{\mathcal{S}}_i$ and $\bm{s}_T = E(\bm{x}_2) \in \bm{\mathcal{S}}_j$). 
$\bm{s}_1$ and $\bm{s}_T$ are linearly interpolated ($\{ \pmb{s}_1, \ldots, \pmb{s}_T \}$) and the intermediate points are used to generate the new images 
$\{ G(\pmb{x}, \pmb{s}_1), \ldots, G(\pmb{x}, \pmb{s}_T) \}$, with $T=20$.
The FID scores are computed by averaging over all the $T \times 1,000$ generated images.
Concerning LPIPS, for each source image
we sample 10 
style codes in each target domain, we generate the corresponding images (without interpolations), and then we compute the LPIPS distances between every pair of images in the same domain, averaging the results across the dataset. 

Since this evaluation method is based on interpolations, for fair comparison we also check the quality of images with FID computed only on some random points in the latent space, as done in \cite{choi2018stargan, choi2019stargan}.  On CelebA-HQ, (Gender translations), we have: StarGAN-v2 \cite{choi2019stargan}, 23.9 and ours: 24.8
(StarGAN-v2 is slightly better than ours). On AFHQ, TUNIT \citep{baek2020tunit}, 17.13; ours, 16.65 (ours is slightly better than \cite{baek2020tunit}). These results  show that, overall, our method does not reduce the image quality of the original  translation task.  Note that the LIPIPS scores reported in  all the tables (of the main manuscript) were computed without interpolations, and they show that our method, in most of the cases, can significantly increase the diversity of the original  translation task.

 In the \noindent\textbf{CelebA-HQ dataset},  we compare also with InterFaceGAN~\citep{shen2020interpreting},
based on StyleGAN~\citep{karras2019style,karras2020analyzing} and trained with high-resolution images. 
However, InterFaceGAN is not designed for MMUIT tasks, and does not have an image encoder.
InterFaceGAN performs face editing by ``moving'' a latent code on the pretrained StyleGAN face representation space along a given direction (e.g. more smile - less smile).
Thus, given a generated image $\pmb{x} = G(\pmb{z})$, where $\pmb{z}$ is a StyleGAN latent code, InterFaceGAN edits $\pmb{z}$ through:
\begin{equation*}
    \pmb{z}' = \pmb{z} + \alpha \pmb{n},
\end{equation*}
where $\pmb{n}$ is the unit normal vector defining a domain-separation  hyperplane  (e.g. smile vs non-smile) and $\alpha$ controls how much positive (or negative) the editing should be (e.g. more smile or less smile). We refer  to~\cite{shen2020interpreting} for additional details.

For I2I translations with InterFaceGAN, we need to use an encoder from images to the StyleGAN face representation space (e.g. \cite{richardson2020encoding, zhu2019lia}). However, the chosen encoder may influence the translation performance. 
To have a fair comparison between InterFaceGAN and MMUIT models in CelebA-HQ, we instead  choose the two reference images  $\pmb{x}_{1}$ and $\pmb{x}_{2}$ (see the  main paper, sec. 6.2), {\em obtained using InterFaceGAN} as follows.
Following~\cite{shen2020interpreting}, for each StyleGAN generated image $\pmb{x}$, we generate $\pmb{x}_{1} = G(\pmb{z} + \alpha_1\pmb{n}))$ and $\pmb{x}_{2} = G(\pmb{z} + \alpha_2\pmb{n})$  with: $\alpha_1 = -3$ (e.g. no smile) and $\alpha_2 = 3$ (e.g. big smile).
These two reference images $\pmb{x}_{1}$ and $\pmb{x}_{2}$ are used for computing the interpolations as described in Sec. 6.2 of the main paper, 
and we emphasize that they are used for all the methods, including
 HomoGAN~\citep{chen2019homomorphic}, StarGAN v2~\citep{choi2019stargan} and ours. 
For the quantitative analysis, we repeat this process 1,000 times, using a different pair $(\pmb{x}_{1}, \pmb{x}_{2})$ at each iteration.

Note that this evaluation protocol does not use any image that is present in the training set of CelebA-HQ. Note also  that the selection of the reference images using InterFaceGAN most likely  helps to increase the InterFaceGAN performance being biased on the  StyleGAN representation space.

In the \noindent\textbf{AFHQ dataset}, we do not  compare with InterFaceGAN, being InterFaceGAN and StyleGAN  not trained on AFHQ.
For this reason, 
both in the MMUIT and the TUNIT settings, the two  reference images $\pmb{x}_{1}$ and $\pmb{x}_{2}$ are simply randomly selected among the real images of the testing 
AFHQ split.



\subsection{Datasets}
We follow the setting in \cite{choi2019stargan} when evaluating the performances on the CelebA-HQ~\citep{karras2017progressive} and the AFHQ dataset~\citep{choi2019stargan}. CelebA-HQ is a High-Quality version of the CelebA~\citep{liu2015deep} dataset, consisting of 30,000 images with a  1024$\times$1024 resolution. We use the  training and the testing lists provided in \cite{choi2019stargan}.
Differently from \cite{choi2019stargan}, we also use the smile attribute for testing. The AFHQ dataset consists of 15,000 high-quality images at 512$\times$512 resolution. It includes three domains ``cat'', ``dog'', and ``wildlife'', each composed of 5,000 images. For each domain, we use the the training and testing lists in \cite{choi2019stargan}. In the MMUIT setting, the CelebA-HQ and the AFHQ datasets are tested with a 256$\times$256 resolution. In the TUNIT setting, following~\cite{baek2020tunit},
we used  test images at a 128$\times$128 resolution.

\subsection{Baselines} 
We use the official and public source codes for all the compared methods, namely StarGAN v2~\citep{choi2019stargan}\footnote[1]{https://github.com/clovaai/stargan-v2}, HomoGAN~\citep{chen2019homomorphic}\footnote[2]{https://github.com/yingcong/HomoInterpGAN}, InterFaceGAN~\citep{shen2020interpreting}\footnote[3]{https://github.com/genforce/interfacegan} and TUNIT~\citep{baek2020tunit}\footnote[4]{https://github.com/clovaai/tunit}. 
Each model
is trained using its  own best hyperparameter values, as selected by the respective authors and provided jointly with the public code.

\section{Additional Results}
\label{AdditionalQualitative}

\begin{figure}[!ht]%
    \centering%
    \includegraphics[width=0.7\columnwidth]{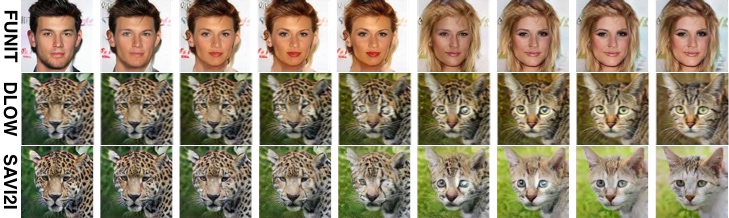}%
    \caption{Qualitative comparison with DLOW \citep{gong2019dlow}, FUNIT \citep{liu2019few} and SAVI2I \citep{mao2022continuous}. This figure shows that obtaining smooth interpolations is a widespread issue.}
    \label{concurrent}%
\end{figure}

\noindent\textbf{Additional comparisons with sota.} The smoothness problem in MMUIT methods is an issue attracting a growing interest in the community, as witnessed, e.g., by \cite{mao2022continuous}, which treats the same problem addressed in our paper. \cref{concurrent} shows three interpolation results taken from Fig. 5 and 6 of \cite{mao2022continuous}, obtained with three different MMUIT methods. 
This figure shows that the non-smoothness problem is shared by other MMUIT models, including SAVI2I, the solution proposed in \cite{mao2022continuous} (which is, by the way, much more complex than our regularization method). Note also that FUNIT \citep{liu2019few}, despite not producing inter-domain artifacts, generates abrupt changes.

\begin{figure*}[!ht]
	\renewcommand{\tabcolsep}{1pt}
	\renewcommand{\arraystretch}{0.8}
	\centering
	\footnotesize
	\begin{tabular}{ccc}
		\includegraphics[width=\linewidth]{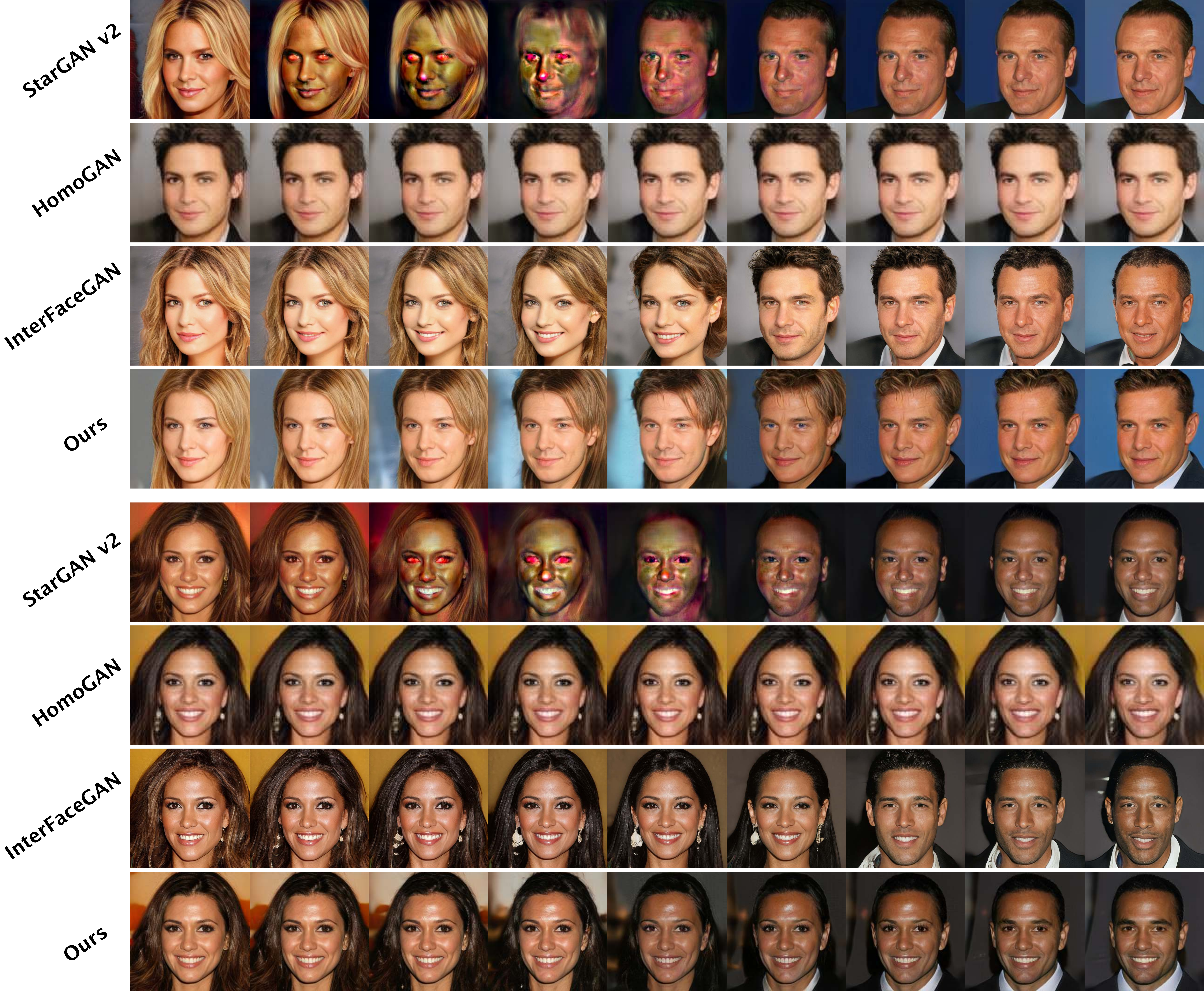} \\
	\end{tabular}
	\caption{CelebA-HQ dataset: qualitative comparisons between  StarGAN v2~\citep{choi2019stargan}, HomoGAN~\citep{chen2019homomorphic}, InterFaceGAN~\citep{shen2020interpreting} and our proposed method on gender translation.
	}
	\label{Fig:celebahq1}
\end{figure*}

\begin{figure*}[!ht]
	\renewcommand{\tabcolsep}{1pt}
	\renewcommand{\arraystretch}{0.8}
	\centering
	\footnotesize
	\begin{tabular}{ccc}
		\includegraphics[width=\linewidth]{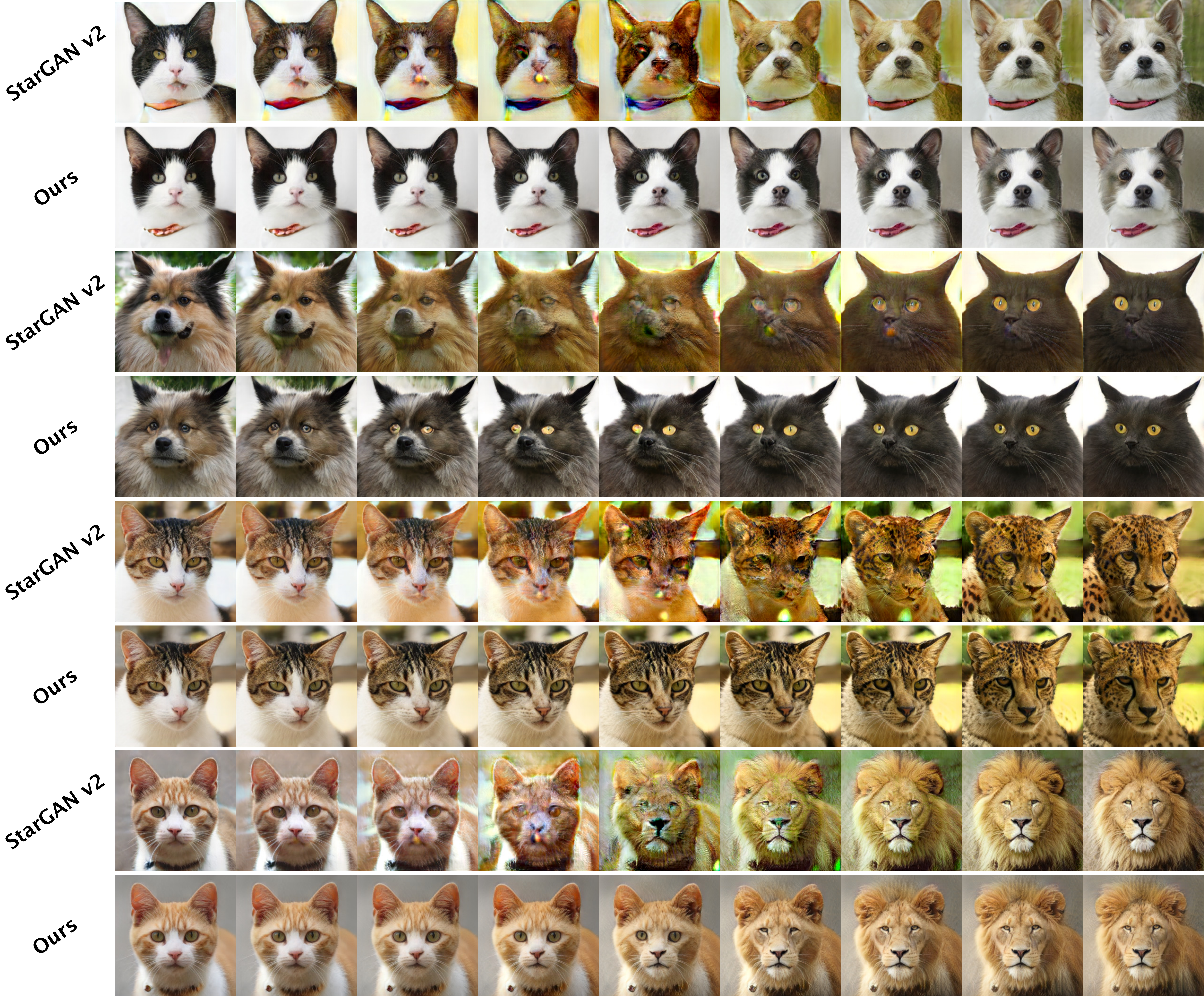} \\
	\end{tabular}
	\caption{AFHQ dataset: qualitative comparisons between  StarGAN v2~\citep{choi2019stargan} and our proposed method on animal translation.
	}
	\label{Fig:afhq1}
\end{figure*}

\begin{figure*}[!ht]
	\renewcommand{\tabcolsep}{1pt}
	\renewcommand{\arraystretch}{0.8}
	\centering
	\footnotesize
	\begin{tabular}{ccc}
		\includegraphics[width=\linewidth]{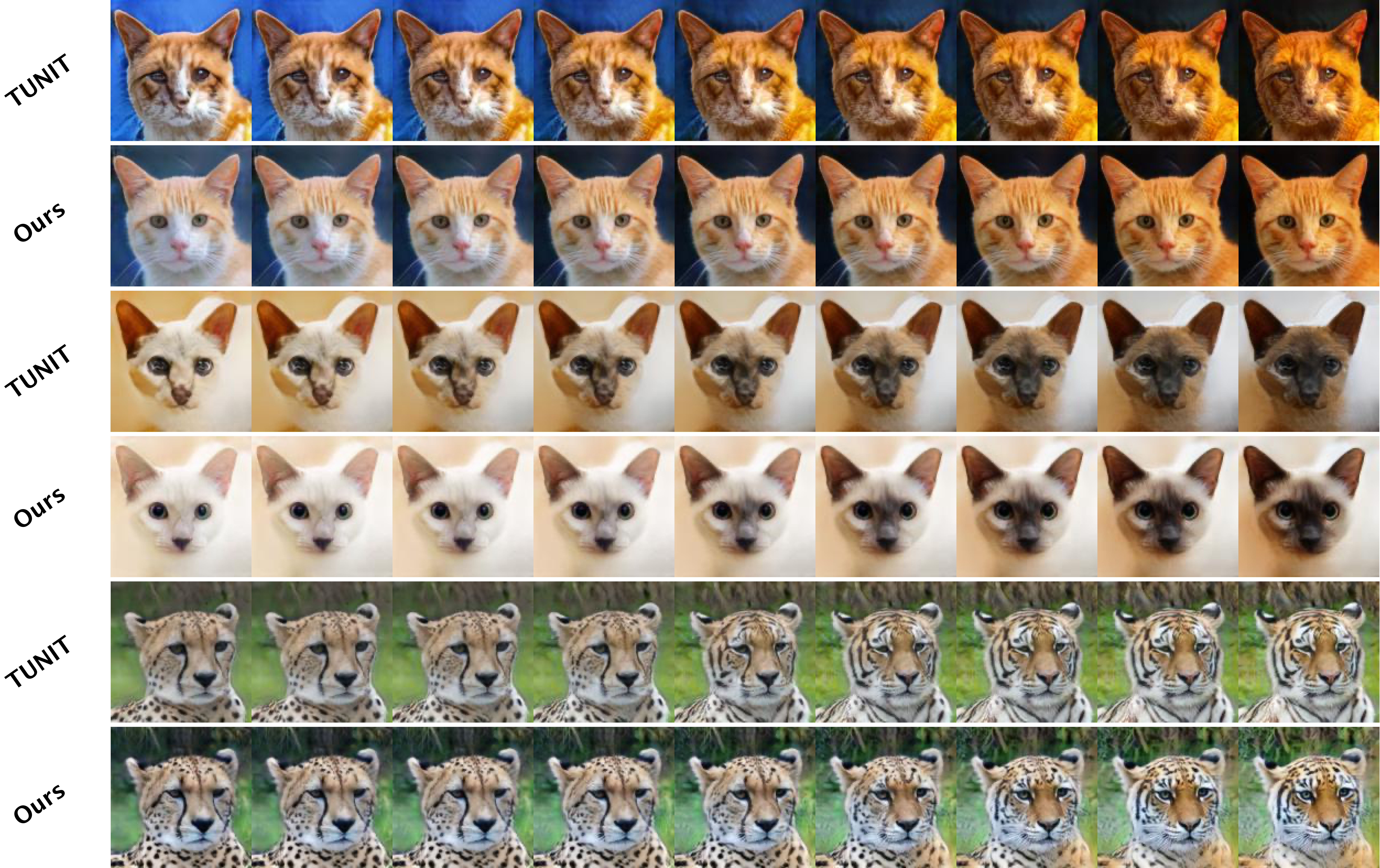} \\
	\end{tabular}
	\caption{AFHQ dataset: qualitative comparisons between  TUNIT~\citep{baek2020tunit} and our proposed method on animal translation.
	}
	\label{Fig:tunit1}
\end{figure*}

\noindent\textbf{Inter-domain Interpolations.}
We show additional qualitative comparisons between different MMUIT state-of-the-art methods and our proposal in \Cref{Fig:celebahq1} and \Cref{Fig:afhq1} for the CelebA-HQ and the AFHQ dataset, respectively. 
In \cref{Fig:tunit1}, we show qualitative comparisons in the TUNIT setting.

Similarly to the results showed in the main paper, we observe that our method generates very smooth inter-domain interpolations, while StarGAN v2 generates artifacts along the interpolation line, and HomoGAN produces very little changes between domains.
In CelebA-HQ, our visual results are very similar to InterFaceGAN, which is based on the training-expensive model StyleGAN~\citep{karras2019style,karras2020analyzing}.

\begin{figure*}[!ht]
	\renewcommand{\tabcolsep}{1pt}
	\renewcommand{\arraystretch}{0.8}
	\centering
	\footnotesize
	\begin{tabular}{ccc}
		\includegraphics[width=\linewidth]{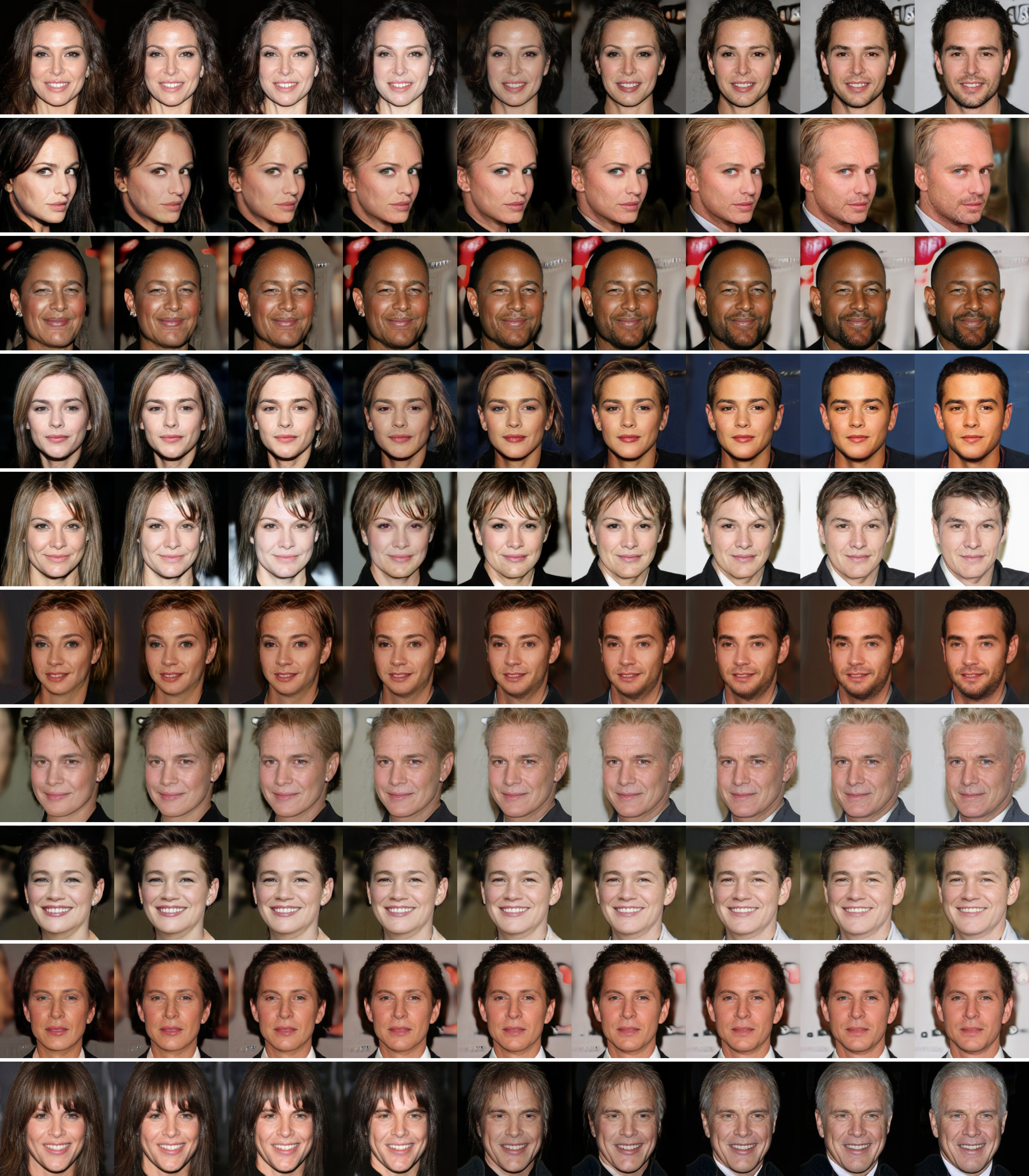} \\
	\end{tabular}
	\caption{More examples of gender translation on the CelebA-HQ dataset~\citep{karras2017progressive}.
	}
	\label{Fig:celebahq-sm1}
\end{figure*}

\begin{figure*}[!ht]
	\renewcommand{\tabcolsep}{1pt}
	\renewcommand{\arraystretch}{0.8}
	\centering
	\footnotesize
	\begin{tabular}{ccc}
		\includegraphics[width=\linewidth]{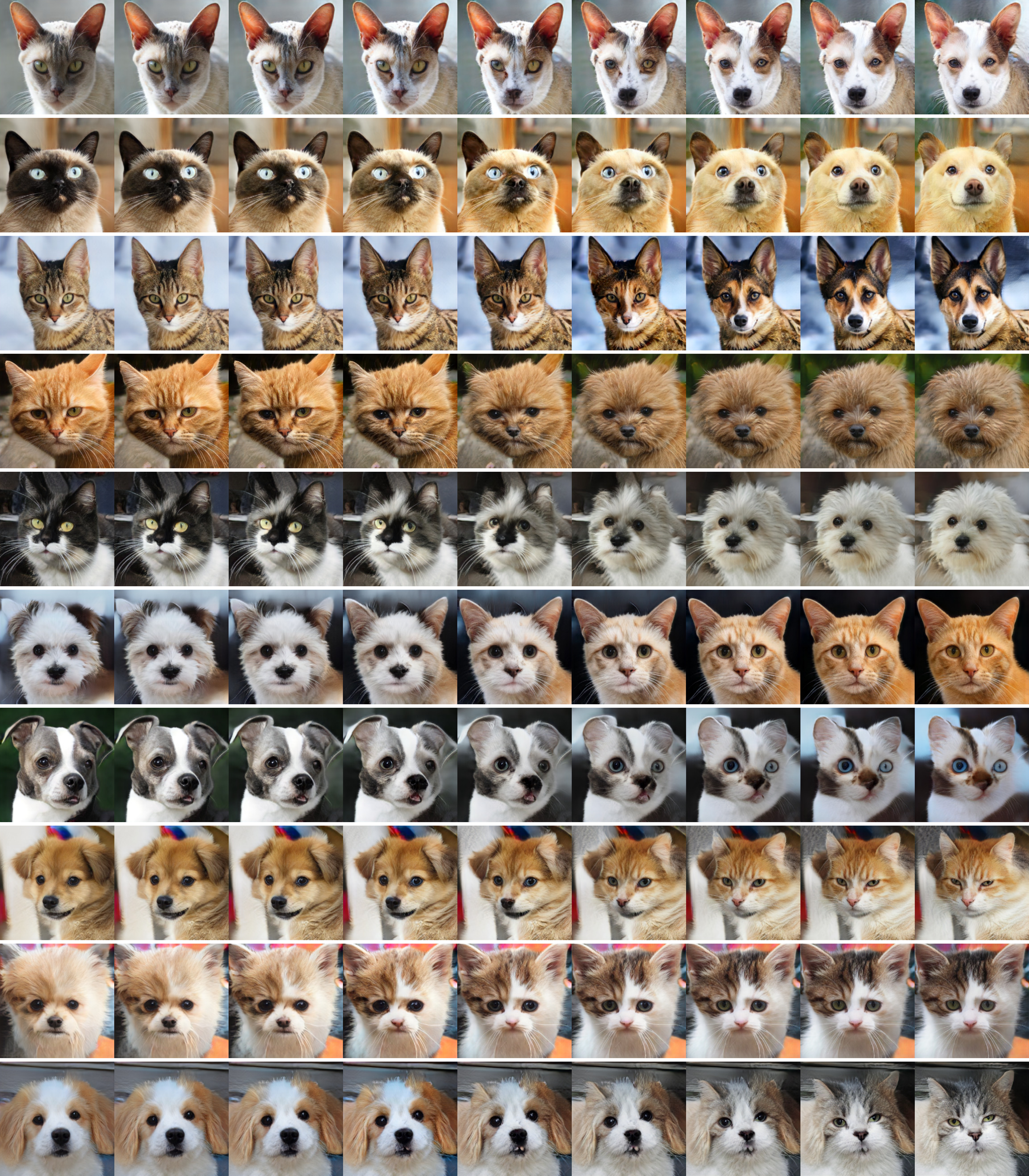} \\
	\end{tabular}
	\caption{More examples of animal face translation on the AFHQ dataset~\citep{choi2019stargan}.
	}
	\label{Fig:afhq2}
\end{figure*}

\Cref{Fig:celebahq-sm1} and \Cref{Fig:celebahq-sm2} show additional qualitative examples on the CelebA-HQ dataset, while   \Cref{Fig:afhq2} and \Cref{Fig:afhq3}
show additional examples on the AFHQ dataset.



\begin{figure*}[!ht]
	\renewcommand{\tabcolsep}{1pt}
	\renewcommand{\arraystretch}{0.8}
	\centering
	\footnotesize
	\begin{tabular}{ccc}
		\includegraphics[width=\linewidth]{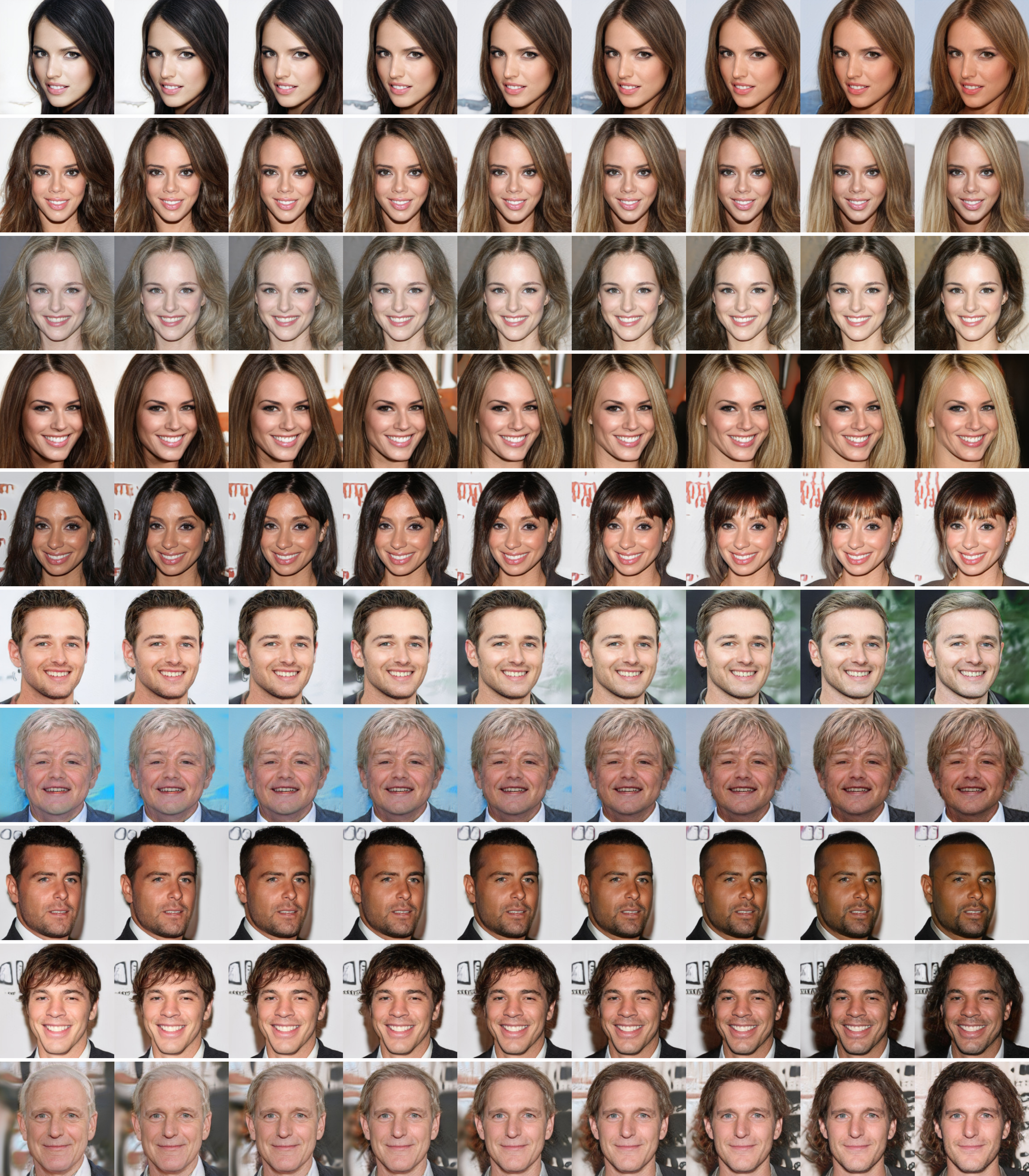} \\
	\end{tabular}
	\caption{Intra-domain interpolation examples of our model on the CelebA-HQ dataset.
	}
	\label{Fig:celebahq-intra-domain}
\end{figure*}

\begin{figure*}[!ht]
	\renewcommand{\tabcolsep}{1pt}
	\renewcommand{\arraystretch}{0.8}
	\centering
	\footnotesize
	\begin{tabular}{ccc}
		\includegraphics[width=\linewidth]{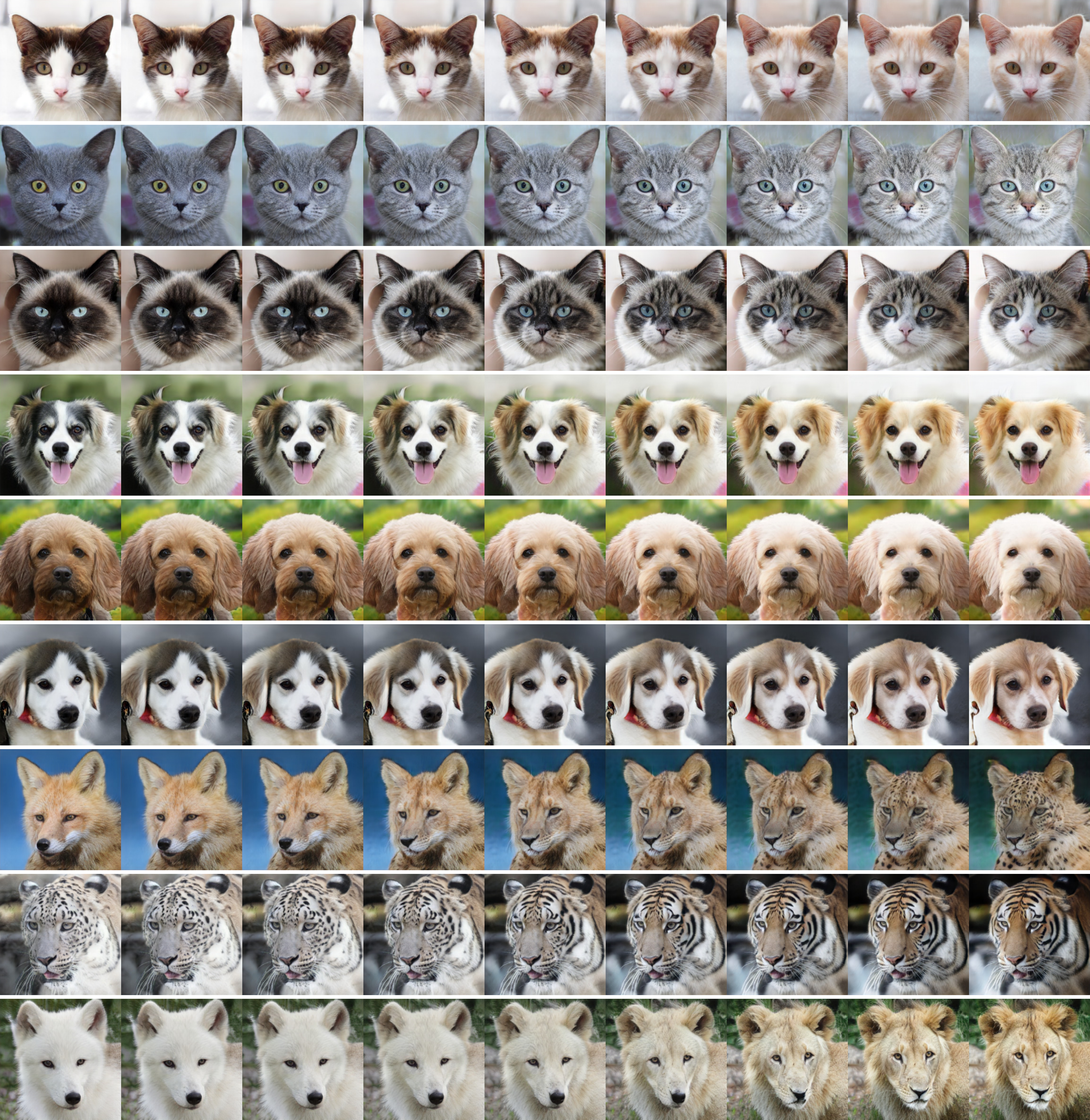} \\
	\end{tabular}
	\caption{Intra-domain interpolation examples of our model on the AFHQ dataset. Note that the ``wildlife'' domain in AFHQ contains different animal species, and this is why, e.g., in the last row, a wolf is transformed into a lion.
	}
	\label{Fig:afhq-intra-domain}
\end{figure*}

\noindent\textbf{Intra-domain Interpolations}
\cref{Fig:celebahq-intra-domain} and \cref{Fig:afhq-intra-domain} show   intra-domain interpolation examples of our model in the MMUIT setting. 

\end{document}